%% file: sample-sigconf.tex
\documentclass[sigconf,nonacm,balance]{acmart}
\usepackage{courier}
\usepackage{colortbl}
\usepackage[nointegrals]{wasysym}
\usepackage{amsthm}
\usepackage{color}
\usepackage{mathtools}
\usepackage{bm}
\usepackage{float}
\usepackage{natbib}
\usepackage{caption}
\usepackage{algorithm}
\usepackage{algorithmic}
\usepackage{tcolorbox}
\usepackage{hhline}
\usepackage{subfiles}
\usepackage{booktabs}
\usepackage{multirow}
\usepackage{stfloats}
\usepackage{lipsum}
\usepackage{arydshln}
\usepackage{graphicx}
\usepackage{subcaption}
\usepackage[normalem]{ulem}

\usepackage{amssymb}
\usepackage{tikz}
\usepackage{amsmath}
\usepackage{enumitem}
\usepackage{xcolor}

\definecolor{mygreen}{rgb}{0.1, 0.6, 0.1}
\definecolor{myblue}{RGB}{0, 112, 192}
\definecolor{myred}{RGB}{255, 0, 0}
\newcommand{\greenbox}[1]{
    \begin{tikzpicture}[baseline=(X.base)]
        \node[fill=green!20, rounded corners=2pt, inner sep=1pt] (X) {#1};
    \end{tikzpicture}
}

\newcommand{\redbox}[1]{
    \begin{tikzpicture}[baseline=(X.base)]
        \node[fill=red!20, rounded corners=2pt, inner sep=1pt] (X) {#1};
    \end{tikzpicture}
}

\definecolor{darkspringgreen}{rgb}{0.09, 0.45, 0.27}
\newcommand{\changed}[1]{\textcolor{black}{#1}}

\AtBeginDocument{%
  }

\copyrightyear{2025}
\acmYear{2025}
\setcopyright{cc}
\setcctype{by}
\acmConference[KDD '25]{Proceedings of the 31st ACM SIGKDD Conference on
Knowledge Discovery and Data Mining V.2}{August 3--7, 2025}{Toronto, ON,
Canada}
\acmBooktitle{Proceedings of the 31st ACM SIGKDD Conference on Knowledge
Discovery and Data Mining V.2 (KDD '25), August 3--7, 2025, Toronto, ON,
Canada}
\acmDOI{10.1145/3711896.3737060}
\acmISBN{979-8-4007-1454-2/2025/08}

\begin{document}

\title{Multi-level Mixture of Experts for Multimodal Entity Linking}
\author{Zhiwei Hu}
\affiliation{
  \institution{School of Computer and Information Technology \\ Shanxi University}
  \city{Taiyuan}
  \country{China}
}
\email{zhiweihu@whu.edu.cn}

\author{Víctor Gutiérrez-Basulto, \,\,  Zhiliang Xiang}
\affiliation{
  \institution{School of Computer Science and Informatics\\ Cardiff University}
  \city{Cardiff}
  \country{UK}
}
\email{{gutierrezbasultov, xiangz6}@cardiff.ac.uk}

\author{Ru Li}
\authornotemark[1]
\affiliation{
  \institution{School of Computer and Information Technology \\ Shanxi University}
  \city{Taiyuan}
  \country{China}
}
\email{liru@sxu.edu.cn}

\author{Jeff Z. Pan}
\authornote{Contact Authors.}
\affiliation{
  \institution{ILCC, School of Informatics\\ University of Edinburgh}
  \city{Edinburgh}
  \country{UK}
}
\email{http://knowledge-representation.org/j.z.pan/}

\subfile{sections/abstract}





\maketitle

\section{Introduction}
\subfile{sections/introduction}

\section{Related Work}
\subfile{sections/related_work}

\section{Method}
\subfile{sections/method}

\section{Experiments}
\subfile{sections/experiment}

\section{Conclusions}
\subfile{sections/conclusion}

\bibliographystyle{ACM-Reference-Format}
\bibliography{sample-base}
\balance
\clearpage
\section*{Appendix}

\subfile{sections/appendix}

\end{document}

%% file: sections/abstract.tex
\begin{abstract}

{Multimodal Entity Linking (MEL) aims to link ambiguous mentions within multimodal contexts to associated entities in a multimodal knowledge base. Existing approaches to MEL  introduce multimodal interaction and fusion mechanisms to bridge the modality gap and enable multi-grained semantic matching. However, they do not address two important problems: ($i$) \textit{mention ambiguity}, \textit{i.e.,} the lack of semantic content caused by the brevity and omission of key information in the mention's textual context; ($ii$) \textit{dynamic selection of modal content}, \textit{i.e.,} to dynamically  distinguish the importance of different parts  of  modal information. To mitigate these issues, we propose a \textbf{M}ulti-level \textbf{M}ixture \textbf{o}f \textbf{E}xperts (\textbf{MMoE}) model for MEL. MMoE has four components: ($i$) the \textit{description-aware mention enhancement} module leverages large language models to identify the WikiData descriptions that best match a mention,  considering  the mention's textual context; ($ii$) the \textit{multimodal feature extraction} module adopts multimodal feature encoders to obtain textual and visual embeddings for both mentions and entities; ($iii$)-($iv$)  the \textit{intra-level mixture of experts} and  \textit{inter-level mixture of experts} modules apply a switch  mixture of experts mechanism to dynamically and adaptively select  features from relevant regions of information. Extensive experiments demonstrate the outstanding performance of MMoE compared to the state-of-the-art.} MMoE's code is available at: \textcolor{blue}{\url{https://github.com/zhiweihu1103/MEL-MMoE}}.

\end{abstract}

%% file: sections/introduction.tex
\begin{figure}[t!]
    \centering
    \includegraphics[width=0.48\textwidth]{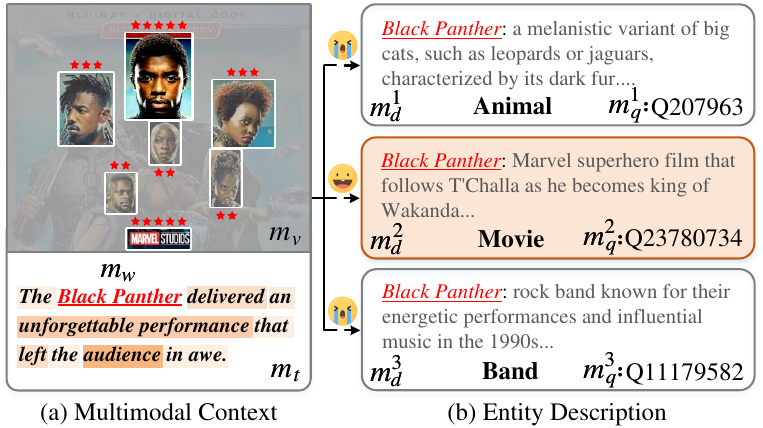}
    \caption{{(a) The given mention \texttt{Black Panther} within a multimodal context. For mention textual context $m_t$, different color depths indicate the importance for the prediction of the mention $m_w$. (b) Wikidata descriptions of entities called \texttt{Black Panther}. The number of $\bigstar$ in $m_v$ indicates the importance level of the corresponding region.}}
    \label{figure_instance}
\end{figure}
\changed{\emph{Entity linking} (EL)~\citep{Tom_2022} is the task of recognising mentions of entities in unstructured textual content and linking them to the corresponding entities in a knowledge graph (KG)~\citep{PVGW2017,PCEH+2017}. EL  supports various downstream information retrieval tasks, such as question answering~\citep{Wenhan_2019,FPKW2019,Shayne_2021,HZLV+2025}, semantic search~\citep{conf/owled/ThomasPS07,reducing2009,Ilaria_2013, Bowei_2023,HZLV+2025}, dialog systems~\citep{Lizi_2018, Ali_2019,WHWX+2025}, and so on~\citep{CLPHC2018,PZSv+2019,Zhiwei_2024_2}. With the increasing availability of  multimodal content, \emph{multimodal entity linking} (MEL), which aims to link ambiguous mentions in multimodal contexts to entities in a multimodal KG (MMKG),  has garnered increasing attention. 
Figure~\ref{figure_instance} shows that  to successfully link the mention \texttt{Black Panther} to its corresponding entities \{$m_q^1, m_q^2, m_q^3$\}, a comprehensive understanding of the associated  textual  $m_t$  and visual  $m_v$ information is required.}
%
\changed{Several approaches have been introduced to handle the MEL task, including ($i$): \textit{vision-and-language pre-trained}~\citep{Alec_2021, Wonjae_2021, Junnan_2021, Zi_2022}, ($ii$): \textit{generative-based}~\citep{OpenAI_2023, Senbao_2024, Xinwei_24}, and ($iii$): \textit{multimodal interaction-based}~\citep{Seungwhan_2018, Omar_2020, Qiushuo_2022, Peng_2022, Pengfei_2023, Pengfei_2024, Zhiwei_2024, Xuhui_2024, Zefeng_2024}. These frameworks have made steady progress, obtaining performance improvements  in various benchmarks like WikiMEL~\citep{Peng_2022}, RichpediaMEL~\citep{Peng_2022}, and WikiDiverse~\citep{Xuwu_2022}.} {However, each of these approaches have their own disadvantages: vision-and-language pre-trained based methods often neglect the fine-grained interactions within modality-specific information, resulting in an inability to fully uncover the latent relationships among the internal elements of a modality. Generative-based methods face challenges in associating visual inputs with their internal entity knowledge, making it difficult to establish connections between fine-grained visual patterns and the corresponding entities. Although the prevailing approaches are based on multimodal interaction, they still suffer from two crucial issues:}

\smallskip
\noindent\textbf{$\triangleright$ Mention Ambiguity.} 
{The context in which textual mentions occur is frequently concise and may incorporate abbreviations, idiomatic expressions, or domain-specific jargon. This scarcity and ambiguity of contextual information can render the accurate identification of the referenced entity infeasible, directly impacting the  performance of a model. Figure~\ref{figure_instance} shows an intuitive example in which  for the mention \texttt{Black Panther}, it is difficult to determine whether \texttt{Black Panther} is an \textit{animal}, a \textit{movie} or a \textit{band} simply based on its given textual context $m_t$. Indeed, $m_t$ might relate to the film's excellence or the actors' outstanding performance, the animal's striking movements, or the band's impressive live performance. }


\smallskip
\noindent\textbf{$\triangleright$ Dynamic Selection of Modal Content.} 
{Existing methods typically encode the full textual sequence or visual image into a single fixed-length vector for modal interaction, ignoring  that different sections within a modality contribute differently to the mention prediction of the corresponding entity. Figure~\ref{figure_instance} presents a typical example illustrating the variation in the significance of different regions within the image and sections of the textual sequence. For instance, for the textual context $m_t$, compared to the definite article \texttt{the}, the phrase \texttt{unforgettable performance} exerts a greater impact on the interpretation of the mention \texttt{Black Panther}. Similarly, for the visual modality knowledge $m_v$, the regions corresponding to actor \texttt{Chadwick Aaron Boseman} and \texttt{Marvel Studios} should receive more attention compared to the region of actor \texttt{Danai Jekesai Gurira}. Thus, it is paramount to dynamically distinguish the importance of different parts of modal information. }


{To address these  two problems, we propose a  \textbf{M}ulti-level \textbf{M}ixture \textbf{o}f \textbf{E}xperts (\textbf{MMoE}) model for the MEL task, composed of the following four components. The \textit{description-aware mention enhancement} module leverages large language models to identify the WikiData descriptions that best match a mention, to compensate for the limitation of concise mentions. Then, the \textit{multimodal feature extraction} module utilizes a pre-trained CLIP model to represent textual tokens and visual patches for both mentions and entities. Afterwards, we design a \textit{switch mixture of experts} (SMoE) mechanism to  dynamically and adaptively select  features from relevant regions of information. Specifically, we introduce SMoE for both intra and inter modality interaction.} 

In summary, our main contributions are three-fold:
\begin{itemize}[itemsep=0.5ex, leftmargin=5mm]
\item We propose the MMoE framework  for MEL which selects the optimal expert network from  the intra- and inter-modality  perspectives, and  dynamically and adaptively selects  features from relevant regions of information.
\item We investigate the issue of mention ambiguity  and use the relevant WikiData description of an entity (identified using a large language model) to enhance the semantic richness of the mention context.
\item We conduct empirical and ablation experiments on three widely-used benchmarks, \textit{i.e.,} WikiMEL, RichpediaMEL and WikiDiverse, showing the effectiveness and generality of MMoE over  existing state-of-the-art models.
\end{itemize}

%% file: sections/related_work.tex
\smallskip
\noindent\textbf{$\triangleright$ Text-based Entity Linking.} 
{Text-based Entity Linking  (EL) focuses on identifying mentions in text and linking them to corresponding entities in knowledge graphs. Approaches to EL can be categorized in \textit{retrieval-based methods} and \textit{generative-based methods}. ($i$) \textit{retrieval-based methods} leverage retrieval mechanisms to filter the most relevant entities~\citep{Ledell_2020, Phong_2018, Matthew_2019, Chenwei_2018, Yixin_2018}. ($ii$) \textit{generative-based methods} utilize the intrinsic knowledge within language models to generate entities associated with mentions~\citep{Jacob_2019, Yinhan_2019, Nicola_2021, Nicola_2022}. While text-based entity linking methods have achieved significant advancements, they cannot be directly applied to the multimodal setting.}

\smallskip
\noindent\textbf{$\triangleright$ Multimodal Entity Linking.}
{Multimodal entity linking (MEL) is an extension of the text-based entity linking task, which utilizes both textual and visual modal knowledge to deal with the ambiguity of entities. Existing MEL methods can be  classified into three categories: \textit{vision-and-language pre-trained methods}, \textit{generative-based methods}, and \textit{multimodal interaction based methods}. ($i$) \textit{vision-and-language pre-trained methods} use large-scale multimodal data to  learn semantic alignment and complementary visual and language knowledge, thereby enhancing its understanding and generation capabilities for the MEL task~\citep{Alec_2021, Wonjae_2021, Junnan_2021, Zi_2022}. ($ii$) \textit{generative-based methods} leverage the strong generative capabilities of large language models and the constrained decoding strategy to ensure the validity of the generated entities~\citep{OpenAI_2023, Senbao_2024, Qi_2024, Xinwei_24}. ($iii$) \textit{multimodal interaction-based methods} employ various mechanisms, such as concatenation~\citep{Omar_2020, Li_2021},  attention~\citep{Seungwhan_2018, Qiushuo_2022, Peng_2022, Dongjie_2022, Gongrui_2023}, matching networks~\citep{Pengfei_2023, Pengfei_2024, Zhiwei_2024, Shezheng_2024} and optimal transport~\citep{Zefeng_2024}, to fusion different modal knowledge from entities and mentions. However, all of these methods still suffer from mention ambiguity and dynamic selection content issues. }

%% file: sections/method.tex
\begin{figure*}[!htp]
    \centering
    \includegraphics[width=1\textwidth]{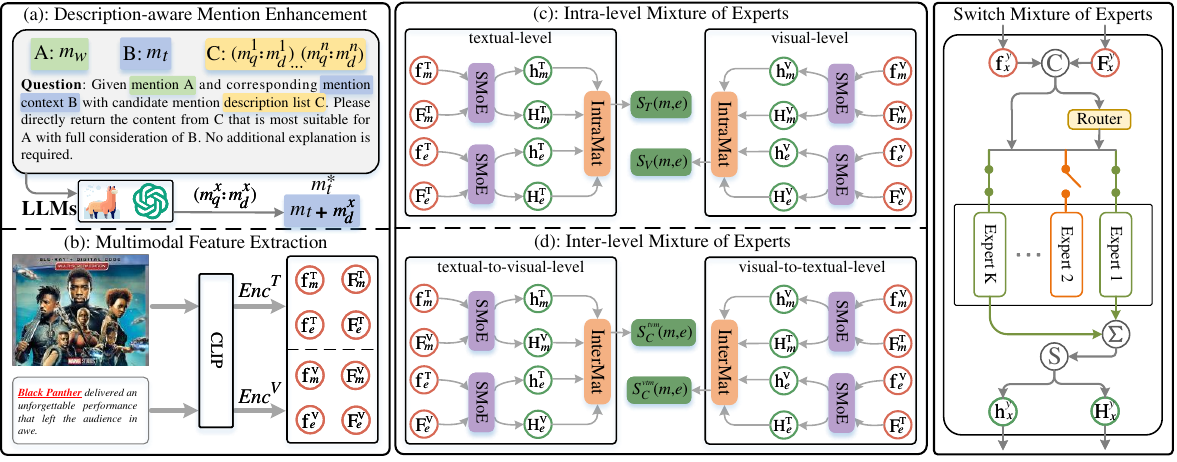}
    \caption{\changed{The structure of the MMoE model, containing four components: (\textit{a}): Description-aware Mention Enhancement (DME), (\textit{b}): Multimodal Feature Extraction (MFE), (\textit{c}): Intra-level Mixture of Experts (IntraMoE) and (\textit{d}): Inter-level Mixture of Experts (InterMoE). IntraMat and InterMat are short for Intra-level Feature Matching and Inter-level Feature Matching modules.}}
    \label{figure_model}
\end{figure*}

In this section, we introduce the MEL problem definition (\S\ref{sec:def}) and  the four components of our MMoE (\S\ref{section_3_2}-\S\ref{section_3_5}). 


\subsection{Problem Definition}
\label{sec:def}
\changed{Intuitively, given a mention $m$ within a multimodal context, the \textit{multimodal entity linking} (MEL) task aims to retrieve the ground truth entity $e$ from the set of entities $\mathcal{E}$ of a knowledge base that is the most relevant to $m$. Each entity  $e\triangleq(e_n, e_a, e_v)$ is composed of an entity name $e_n$, textual attribute $e_a$ and a visual image $e_v$. Each mention is defined as a triple $m\triangleq(m_w, m_t, m_v)$, with  $m_w$ a mention word,  $m_t$ a textual context, and  $m_v$ a visual image.
Formally,  MEL can be formulated as maximizing the log-likelihood over the training set $\mathcal{D}$ while optimizing the model parameters $\theta$ as follows:
\begin{equation}
\label{equation_1}
\theta^*=\underset{\theta}{{\rm max}}\sum_{(m, e)\in\mathcal{D}}{\rm log}\,p_{\theta}(e|m, \mathcal{E})
\end{equation}
where $\theta^*$ denotes the final parameters,  and $p_{\theta}(e|m, \mathcal{E})$ is a score function used to calculate the similarity between mention $m$ and the candidate entity $e \in \mathcal{E}$.}


\subsection{Description-aware Mention Enhancement}
\label{section_3_2}
\changed{Textual mentions and associated contexts are frequently concise  and may include abbreviations. For instance, in the WikiDiverse dataset  mention contexts have an average length of 11 words. This brevity can give rise to the ``\emph{mention ambiguity}'' phenomenon. More precisely, a limited number of words  fails to provide sufficient semantic information, complicating the extraction of adequate knowledge from the context surrounding a mention. Figure~\ref{figure_instance} provides a clear illustration of this issue,  where  the mention word $m_w$ is \texttt{Black Panther} and  the mention textual context $m_t$ is \texttt{The Black Panther delivered...}. {Given the succinctness of $m_t$, it is challenging to ascertain whether \texttt{Black Panther} refers to an animal, a movie, or a band, because $m_t$ might relate to the awe-inspiring movements or posture displayed by the animal, the brilliance of the film or the outstanding performance of actors, or the band's outstanding performance and musical impact during the show.} So, such ambiguity negatively impacts the effectiveness of frameworks for the MEL task.}


\changed{However, WikiData~\citep{Denny_2014} provides a description for each entity, which might serve to mitigate the mention ambiguity issue. Nonetheless, a single entity name may correspond to several entities within WikiData, each associated with its own description. For instance, the mention \texttt{Black Panther} in Figure~\ref{figure_instance} can refer to  the WikiData entities \texttt{Q207963}, \texttt{Q23780734}, and \texttt{Q11179582}. Selecting the most appropriate description in a given context is thus a critical challenge to overcome for the succesful use of WikiData information. Recently, large language models (LLMs), such as  ChatGPT~\citep{OpenAI_2023} and LLaMA~\citep{Hugo_2023}, have demonstrated strong capabilities in knowledge storage and semantic evaluation. Motivated by this, we introduce the  LLM-based \textbf{D}escription-aware \textbf{M}ention \textbf{E}nhancement (\textbf{DME}) module, designed to augment the knowledge associated with a mention. 
Specifically, starting with the mention word $m_w$ and mention textual context $m_t$, to generate the description-aware enhanced mention content $m_t^{\ast}$, we proceed through three steps as depicted  in Figure~\ref{figure_model}(a): 
}
\begin{enumerate}[itemsep=0.5ex, leftmargin=4mm]
\item \changed{Given the mention word $m_w$, we  retrieve entities from WikiData that share the same name with $m_w$. The set of all these entities is  denoted as  $M_w$ = $\{(m_q^1,m_d^1)...$$(m_q^i,m_d^i)...$$(m_q^n,m_d^n)\}$, where  $n$ is the number of retrieved entities, and $m_d^i$ represents the description  corresponding to the \textit{i}-th entity with  id $m_q^i$. For instance, as shown in Figure~\ref{figure_instance}, $m_q^1$ is \texttt{Q207963} and the corresponding description $m_d^1$ is \texttt{Black Panther: a melanistic variant...}.}


\item {Given $M_w$, we use the  prompt in Figure~\ref{figure_model}(a), to employ LLMs to identify the entity-description pair that is the most pertinent for the mention $m_w$ and its textual context $m_t$, we denote the most appropriate entity-description pair as $(m_q^x,m_d^x)$. For example, as shown in Figure~\ref{figure_instance}, the most relevant WikiData entity id and description corresponding to mention \texttt{Black Panther} is $m_q^2$ and $m_d^2$.}


\item {We concatenate the mention textual context $m_t$ with the most relevant mention description $m_d^x$ to obtain the description-aware enhanced  mention textual context $m_t^{\ast}=m_t+m_d^x$.}

\end{enumerate}

\subsection{Multimodal Feature Extraction}
\label{section_3_3}
\changed{Given the entity $e$ and mention $m$ with their corresponding textual and visual knowledge, we generate their initial embeddings as done in~\citep{Pengfei_2023, Zhiwei_2024, Zefeng_2024}, as shown in Figure~\ref{figure_model}(b). Specifically, we adopt the contrastive language-image pretraining architecture (CLIP)~\citep{Alec_2021} as the multimodal encoder, which contains a pre-trained BERT~\citep{Jacob_2019} textual encoder $\mathit{Enc^T}$ and a pre-trained ViT~\citep{Alexey_2021} visual encoder $\mathit{Enc^V}$: }
\begin{equation}
\label{equation_2}
\left\{
     \begin{array}{l}
     \textbf{F}_m^T=\mathit{Enc^T}([\texttt{CLS}]\,m_w\,[\texttt{SEP}]\,m_t^\ast) \vspace{1.5mm}\\
     \textbf{F}_e^T=\mathit{Enc^T}([\texttt{CLS}]\,e_n\,[\texttt{SEP}]\,m_a) \vspace{1.5mm}\\
     \textbf{F}_m^V=\mathit{Enc^V}(m_v),\,\textbf{F}_e^V=\mathit{Enc^V}(e_v)\\
     \end{array} \right.
\end{equation}
\changed{{where $\textbf{F}_m^T\in\mathbb{R}^{L_1\times d},\textbf{F}_e^T\in\mathbb{R}^{L_2\times d}$ represent sequential textual features for the mention and entity sentence, $L_1$ and $L_2$ denote the length of the token sequence of the mention and entity, respectively, and $d$ is the dimension of embeddings. $\textbf{F}_m^V\in\mathbb{R}^{P_1\times d},\textbf{F}_e^V\in\mathbb{R}^{P_2\times d}$ denote visual features for the mention and entity patches, $P_1$ and $P_2$ represent the number of patches of the mention and entity, respectively.} In addition to these fine-grained features, we also obtain coarse-grained features $\textbf{f}_m^T,\textbf{f}_e^T,\textbf{f}_m^V,\textbf{f}_e^V\in\mathbb{R}^{d}$ for a  more comprehensive  semantic understanding, which can be achieved by using the embedding of the special token $[\texttt{CLS}]$ for the textual encoder $\mathit{Enc^T}$ and the special token $[\texttt{EOS}]$ for the visual encoder $\mathit{Enc^V}$.}


\subsection{Intra-level Mixture of Experts}
\label{section_3_4}
\changed{Considering that not every part (region/section) of unimodal information contributes equally to the final prediction, different tokens and  patches in the textual and visual information characterize  the entity or mention content to different extents. For example, as shown in Figure~\ref{figure_instance}, for the mention context $m_t$, it is obvious that \texttt{unforgettable performance} should be given higher attention than the definite article \texttt{the}. To help dealing with such redundant noise, we believe specialized screening and learning of unimodal information through sample adaptive mechanisms are paramount. To achieve this vision, we need to address two important questions: (1) \textit{What guides the selection of specific regions within a modality?} (2) \textit{How to dynamically learn the importance of different regions of information in various samples?}}


\smallskip
\noindent\textbf{$\triangleright$ SMoE: Switch Mixture of Experts.} 
\changed{To address these challenges, we introduced the \textbf{S}witch \textbf{M}ixture \textbf{o}f \textbf{E}xperts (\textbf{SMoE}) mechanism to dynamically learn the contribution of different regions of information within a modality. To solve the first problem, we use coarse-grained and fine-grained features to guide the supervision of specific regions. To solve the second problem,  we introduce a mixture of experts mechanism to adaptively and dynamically select information from different regions. Specifically,  SMoE is composed of the following three steps:}

\begin{enumerate}[itemsep=0.5ex, leftmargin=4mm]
\item \textit{ Multi-granular  Fusion of Features.}
\changed{Given the coarse-grained and fine-grained features $\textbf{f}_x^y$, $\textbf{F}_x^y$, we first perform a concatenation operation and then apply multi-layer perceptron (MLP) layers to obtain the fusion feature $\textbf{P}_x^y=\texttt{MLP}([\textbf{F}_x^y;\textbf{f}_x^y])$. This is  simple and intuitive, and  can also achieve a bidirectional effect from coarse-grained to fine-grained and vice versa in subsequent operations.}

\item \textit{Dynamic Selection of Features.} 
\changed{The main components of a SMoE layer are a network $\textbf{\texttt{R}}$ as a sparse router and an expert network $\textbf{\texttt{E}}$. Given a token representation ${p}_x^y$ from $\textbf{P}_x^y$, we process it sparsely using $k$ out of the $K$ available experts to obtain the final representation ${q}_x^y$, which is formally defined as:}
\begin{equation}
\label{equation_3}
\left\{
	\begin{array}{l}
	\textbf{\texttt{R}}({p}_x^y)=\texttt{Softmax}(\textbf{W}_{\textbf{\texttt{R}}}\otimes{p}_x^y) \vspace{1.5mm}\\
    q_x^y=\displaystyle\sum_{i=1}^{k}{\textbf{\texttt{R}}_i\circledast \textbf{\texttt{E}}_i,\,\textbf{\texttt{E}}_i=\texttt{FFN}_i({p}_x^y)}\\
	\end{array} \right.
\end{equation}
\changed{where $\otimes$ and $\circledast$ denote  matrix and element-wise multiplication respectively. $\textbf{W}_{\textbf{\texttt{R}}}$ is a learnable parameter, $\texttt{FFN}_i$ represents a feed-forward layer and its parameters are independent between different experts, $\textbf{\texttt{R}}_i$ denotes the activation value of the \textit{i}-th expert, considering that the acquisition of the $\textbf{\texttt{R}}_i$ value is adaptive, so its behavior is intrinsically dynamic. The final output ${q}_x^y$ of $k$ activated experts are the linearly weighted combination of each expert's output $\textbf{\texttt{E}}_i$ on the token output by the router $\textbf{\texttt{R}}_i$. By performing similar operations on each token we can obtain the representation $\textbf{Q}_x^y$ after multiple layers of SMoE.}


\item \textit{ Multi-granular Split of Features.} 
\changed{Considering that the dynamic selection process in Step (2) is completed by coarse-grained and fine-grained features  $\textbf{f}_x^y$, $\textbf{F}_x^y$, the interaction between $\textbf{f}_x^y$ and $\textbf{F}_x^y$ is realized to a certain extent, and the introduction of the multi-layer MoE  further diversifies such interaction. We further use a split operation based on the dimension to restore $\textbf{Q}_x^y$ to $\textbf{h}_x^y$ and $\textbf{H}_x^y$ to facilitate the subsequent calculation of feature matching scores. This is  formally expressed as $[\textbf{h}_x^y; \textbf{H}_x^y]=\texttt{Split}(\textbf{Q}_x^y)$.}

\end{enumerate}

\smallskip
\noindent\textbf{$\triangleright$ IntraMat: Intra-level Feature Matching.}
\changed{To fully consider the unimodal correlations between mentions and entities, we design the \textbf{Intra}-level Feature \textbf{Mat}ching (\textbf{IntraMat}) module, which is composed of the \textit{Coarse-grained  Matching (CM)} and \textit{Fine-grained  Matching (FM)} sub-modules. As shown in Figure~\ref{figure_model}(c), given the mention $m$ with coarse-grained and fine-grained textual embeddings $\textbf{f}_m^T$, $\textbf{F}_m^T$, and entity $e$ with coarse-grained and fine-grained textual embeddings $\textbf{f}_e^T$, $\textbf{F}_e^T$, we first leverage the SMoE mechanism to obtain the enhanced embeddings $\textbf{h}_m^T$, $\textbf{H}_m^T$ and $\textbf{h}_e^T$, $\textbf{H}_e^T$. Then we integrate two types of matching scores as follows: }

\begin{itemize}[itemsep=0.5ex, leftmargin=3mm]
\item \textit{Coarse-grained  Matching (CM).} \changed{We directly perform the dot product $\odot$ between the mention and entity coarse-grained textual embeddings $\textbf{h}_m^T$ and $\textbf{h}_e^T$ formulated as:}
\begin{equation}
\label{equation_4}
\bm{\mathcal{S}}_T^\texttt{cm}(m,e)=\textbf{h}_e^T\odot\textbf{h}_m^T
\end{equation}
\item \textit{Fine-grained  Matching (FM).} \changed{We establish the fine-grained textual unimodal correlation assignments based on an attention mechanism~\citep{Ashish_2017} as follows:}
\begin{equation}
\label{equation_5}
\left\{
	\begin{array}{l}
    \begin{footnotesize}\textbf{M}=\textbf{H}_e^T\otimes\textbf{W}_q,\textbf{K}=\textbf{H}_m^T\otimes\textbf{W}_k,\textbf{V}=\textbf{H}_m^T\otimes\textbf{W}_v\vspace{1.5mm}\end{footnotesize}\\
    \textbf{A}_{m\rightarrow e}^T=\texttt{Softmax}(\textbf{X}),\,\textbf{X}=\textbf{M}\otimes\textbf{K}/\sqrt{d}\vspace{1.5mm}\\
    \textbf{G}_{m\rightarrow e}^T=\texttt{mean}(\textbf{A}_{m\rightarrow e}^T\otimes\textbf{V})\vspace{1.5mm}\\
    \bm{\mathcal{S}}_T^\texttt{fm}(m,e)=\textbf{h}_e^T\odot\textbf{G}_{m\rightarrow e}^T\vspace{1.5mm}\\
    \bm{\mathcal{S}}_T(m,e)=(\bm{\mathcal{S}}_T^\texttt{cm}(m,e)+\bm{\mathcal{S}}_T^\texttt{fm}(m,e))/2\\
	\end{array} \right.
\end{equation}
\changed{where $\textbf{W}_q$, $\textbf{W}_k$, $\textbf{W}_v$ are learnable parameters, $\textbf{A}_{m\rightarrow e}^T$ is the assigned correlation weight from mention textual tokens embeddings $\textbf{H}_m^T$ to entity textual tokens embeddings $\textbf{H}_e^T$. $\textbf{G}_{m\rightarrow e}^T$ represents the aggregated representation of $\textbf{H}_m^T$ and $\textbf{H}_e^T$ after attention interaction. Afterwards, the intra-level textual matching score $\bm{\mathcal{S}}_T(m,e)$ is defined as the average of the coarse-grained score $\bm{\mathcal{S}}_T^\texttt{cm}(m,e)$ and fine-grained score $\bm{\mathcal{S}}_T^\texttt{fm}(m,e)$. }

\changed{Similarly, the intra-level visual matching score $\bm{\mathcal{S}}_V(m,e)$ is defined as the average of $\bm{\mathcal{S}}_V^\texttt{cm}(m,e)$ and $\bm{\mathcal{S}}_V^\texttt{fm}(m,e)$.}

\end{itemize}

\subsection{Inter-level Mixture of Experts}
\label{section_3_5}
\changed{Relying exclusively on unimodal content often leads to ambiguous interpretations. Consider the example presented in Figure~\ref{figure_instance}, when constrained only to the textual modality $m_t$, it remains uncertain whether the mention \texttt{Black Panther} refers to an \textit{animal} or a \textit{movie}. However, incorporating the visual modality $m_v$ helps to alliviate this ambiguity, substantially increasing the likelihood that \texttt{Black Panther} is identified as a \textit{movie}.  This enhancement stems from the fact that visual information provides spatial structural knowledge absent in the textual representation, while textual information offers semantic details not captured visually. A synergistic integration of both modalities compensates for the deficiencies inherent to each of them, leading to a more robust understanding and interpretation of the content.}


\smallskip
\noindent\textbf{$\triangleright$ InterMat: Inter-level Feature Matching.}
\changed{To  alleviate this content gap between mentions and entities, we  introduce an adaptive multimodal feature interaction mechanism: 
the \textbf{Inter}-level Feature \textbf{Mat}ching (\textbf{InterMat}) module contains the \textit{textual visual matching (TVM)} and \textit{visual textual matching (VTM)} sub-modules. Considering that they have similar operations except for the input, we focus on the  implementation process of TVM. Specifically, as shown in Figure~\ref{figure_model}(d), given the input embeddings \{$\textbf{f}_m^T$, $\textbf{F}_m^V$, $\textbf{f}_e^T$, $\textbf{F}_e^V$\} associated with mention $m$ and entity $e$, similar to IntraMat, we leverage SMoE to obtain enhanced embeddings \{$\textbf{h}_m^T$, $\textbf{H}_m^V$, $\textbf{h}_e^T$, $\textbf{H}_e^V$\}. Then, we apply a gated function based on the extracted embeddings ($\textbf{h}_m^T$, $\textbf{H}_m^V$) as follows:} 
\begin{equation}
\label{equation_6}
\left\{
    \begin{array}{l}
    \textbf{h}_m^{T\star}=\texttt{Tanh}(\textbf{h}_m^T)\vspace{1.5mm}\\
    \textbf{H}_m^{V\star}=\texttt{Softmax}(\textbf{h}_m^T\otimes\textbf{H}_m^V)\otimes\textbf{H}_m^V\vspace{1.5mm}\\
    \textbf{E}_m^\texttt{tvm}=\texttt{LayerNorm}(\textbf{h}_m^{T\star}\circledast\textbf{h}_m^T+\textbf{H}_m^{V\star})\\
    \end{array} \right.
\end{equation}
\changed{ By replacing inputs $\textbf{h}_m^T$ and $\textbf{H}_m^V$ with $\textbf{h}_e^T$ and $\textbf{H}_e^V$ in the  operations in Equation~\ref{equation_6}, we can also obtain the entity-side context embedding $\textbf{E}_e^\texttt{tvm}$, then the score is calculated by the dot product:}
\begin{equation}
\label{equation_7}
\bm{\mathcal{S}}_C^\texttt{tvm}(m,e)=\textbf{E}_e^\texttt{tvm}\odot\textbf{E}_m^\texttt{tvm}
\end{equation}
\changed{Using a similar approach, we can obtain the visual textual matching score $\bm{\mathcal{S}}_C^\texttt{vtm}(m,e)$, and acerage $\bm{\mathcal{S}}_C^\texttt{tvm}(m,e)$ and $\bm{\mathcal{S}}_C^\texttt{vtm}(m,e)$ to get the final matching score $\bm{\mathcal{S}}_C(m,e)$ of the InterMoE module.}

\subsection{Overall Score and Loss Function} \changed{The symmetrical design of the IntraMoE and InterMoE modules introduces comparable features as well as similarity measures from different perspectives. We further collect all matching scores and obtain the overall matching score $\bm{\mathcal{S}}_O(m,e)$. Inspired by~\citep{Pengfei_2023, Zefeng_2024}, we design a contrastive training objective on the different matching scores as follows:  }
\begin{equation}
\label{equation_8}
\left\{
	\begin{array}{l}
    \bm{\mathcal{S}}_O(m,e)=\displaystyle\sum_{X\in\{T,V,C\}}{\bm{\mathcal{S}}_X(m,e)}\vspace{1.5mm}\\
    \bm{\mathcal{L}}_X=-{\rm log}\displaystyle\frac{{\rm exp}(\bm{\mathcal{S}}_X(m,e))}{\displaystyle\sum_{e'\in\mathcal{E}}{{\rm exp}(\bm{\mathcal{S}}_X(m,e'))}}\vspace{1.5mm}\\
    \bm{\mathcal{L}}=\displaystyle\sum_{X\in\{O,T,V,C\}}{\bm{\mathcal{L}}_X}\\
	\end{array} \right.
\end{equation}
\changed{where $\mathcal{E}$ denotes all candidate entities, $e'$ serve as the in-batch negative samples for entity $e$. $\bm{\mathcal{L}}$ is the final loss of our model, which considers the overall loss $\bm{\mathcal{L}}_O$, unimodal losses $\bm{\mathcal{L}}_T$, $\bm{\mathcal{L}}_V$, and multimodal loss $\bm{\mathcal{L}}_C$.}


%% file: sections/experiment.tex

{To evaluate the effectiveness of MMoE, we aim to explore the following four research questions:
\begin{itemize}[itemsep=0.5ex, leftmargin=5mm]
\item 
\textbf{RQ1 (Effectiveness):} How MMoE performs compared to the state-of-the-art under different conditions?
\item 
\textbf{RQ2 (Ablation studies):} How different components contribute to MMoE's performance?
\item 
\textbf{RQ3 (Parameter sensitivity):} How hyper-parameters influence MMoE's performance?
\item
\textbf{RQ4 (Complexity analysis and Case Study):} What is the amount of computation and parameters used by MMoE? We provide additional results and a case study in \textcolor{magenta}{\textbf{\hyperlink{case_study}{\textcolor{blue}{Appendix B}}}} and \textcolor{blue}{\textbf{ \hyperlink{additional_results}{\textcolor{blue}{Appendix C}}}}.
\end{itemize}
}

\subsection{Experimental Setup}
\label{experimental_setup}
\smallskip


\noindent\textbf{$\triangleright$ Datasets.}
{We conduct experiments on three well-known datasets, \textbf{WikiMEL}~\citep{Peng_2022}, \textbf{RichpediaMEL}~\citep{Peng_2022}, and \textbf{WikiDiverse}~\citep{Xuwu_2022}.  The \textbf{WikiMEL} dataset~\citep{Peng_2022} contains more than 22K samples, where entities are collected from WikiData~\citep{Denny_2014} and the corresponding textual and visual  knowledge from WikiPedia;  \textbf{RichpediaMEL}~\citep{Peng_2022} contains more than 17K samples, where entities are gathered from Richpedia~\citep{Meng_2020} and the corresponding multimodal content is obtained from WikiPedia;  \textbf{WikiDiverse}~\citep{Xuwu_2022} contains over 15K samples, where multimodal information is collected from WikiNews and covers various topics. We follow the training, validation, and test set splitting by~\citep{Pengfei_2023, Qi_2024, Zhiwei_2024, Xuhui_2024, Zefeng_2024}. The statistics of three datasets are summarized in Table~\ref{table_statistics_datasets}.}

\begin{table}[!htp]
\setlength{\abovecaptionskip}{0.03cm} 
\renewcommand\arraystretch{1.1}
\setlength{\tabcolsep}{0.6em}
\centering
\small
\begin{tabular*}{\linewidth}{@{}c|ccc@{}}
\hline
\multicolumn{1}{c|}{\textbf{Statistics}} &\multicolumn{1}{c}{\textbf{WikiMEL}} &\multicolumn{1}{c}{\textbf{RichpediaMEL}} &\multicolumn{1}{c}{\textbf{WikiDiverse}}\\
\hline
\# Num. of sentences  &22,070  &17,724  &7,405 \\
\# Num. of entities  &109,976  &160,935  &132,460 \\
\# entities with Img. &67,195  &86,769  &67,309 \\
\# Num. of mentions  &25,846  &17,805  &15,093 \\
\# Img. of mentions   &22,136  &15,853  &6,697 \\
\# mentions in train  &18,092  &12,463  &11,351 \\
\# mentions in valid  &2,585  &1,780  &1,664 \\
\# mentions in test  &5,169  &3,562  &2,078 \\
\hline
\end{tabular*}
\caption{Statistics of three different datasets. "Num." and "Img." denote number and image, respectively.}
\label{table_statistics_datasets}
\end{table}

\smallskip
\noindent\textbf{$\triangleright$ Evaluation Metrics.} \changed{Given the similarity scores between a mention and candidate entities, we sort these scores in descending order to calculate \textbf{MRR} and \textbf{Hits@\textit{n}} (abbreviated sometimes as \textbf{H@\textit{n}}, \textit{n}$\in\{1,3,5\}$). MRR stands for the inverse of the rank for the first correct entity, while Hits@\textit{n} represents the rank for the first correct entity ranked among the top-\textit{n} answers. For all metrics, the larger the values, the better the performance.}

\smallskip
\noindent\textbf{$\triangleright$ Implementation Details.}
\changed{We conduct experiments on four 24G GeForce RTX 4090 GPUs, adopt AdamW~\citep{Ilya_2019} as optimiser, and use the grid search method to select the optimal hyperparameter setting for the MMoE model on different datasets. Consistent with~\citep{Pengfei_2023, Zefeng_2024}, we apply the pre-trained CLIP-ViT-Base-Pathch32 model as the embedding encoder for textual and visual information. For the textual modality, we set up parameter experiments with different max text length. For the visual modality, we first scale the images into $224\times 224$ resolution to conform with the input of the ViT model, then we set the number of patches to $32$. We adopt ``\texttt{gpt-3.5-turbo-0613}'' in the DME module as the LLM to order the description list. We focus on the hyperparameters that have a greater impact on experimental performance, including: \textit{the number of Experts $K$}, \textit{the number of top-experts $k$}, \textit{the embedding dimension $d$}, and \textit{the max text length}, the corresponding search range of candidate values for different hyperparameters are shown in Table~\ref{table_hyperparameters}. }

\begin{table}[!hp]
\renewcommand\arraystretch{1.1}
\setlength{\tabcolsep}{1.5em}
\centering
\small
\begin{tabular*}{0.92\linewidth}{@{}cc@{}}
\hline
\multicolumn{1}{l}{\textbf{Hyperparameters}} & \multicolumn{1}{c}{\textbf{Search Space}} \\
\midrule
\multicolumn{1}{l}{\# Number of Experts $K$}   &\multicolumn{1}{c}{\{$2,\,4,\,6,\,8,\,10$\}} \\
\multicolumn{1}{l}{\# Number of Top-experts $k$}   &\multicolumn{1}{c}{\{$1,\,2,\,3,\,4$\}} \\
\multicolumn{1}{l}{\# Embedding Dimension $d$}   &\multicolumn{1}{c}{\{$48,\,64,\,80,\,96,\,112$\}} \\
\multicolumn{1}{l}{\# Max Text Length}   &\multicolumn{1}{c}{\{$20,\,30,\,40,\,50,\,60$\}} \\
\hline
\end{tabular*}
\caption{The main hyperparameters search space of MMoE model.}
\label{table_hyperparameters}
\end{table}
\vspace{-5mm}

\begin{table*}[!htp]
\setlength{\abovecaptionskip}{0.18cm}
\renewcommand\arraystretch{1.13}
\setlength{\tabcolsep}{0.903em}
\centering
\small
\begin{tabular*}{\linewidth}{@{}ccccccccccccc@{}}
\hline
\multicolumn{1}{c|}{\multirow{1}{*}{\textbf{Methods}}} & \multicolumn{4}{c|}{\textbf{WikiMEL}} & \multicolumn{4}{c|}{\textbf{RichpediaMEL}} & \multicolumn{4}{c}{\textbf{WikiDiverse}}\\
\hline
\multicolumn{1}{c|}{\multirow{1}{*}{\textbf{Metrics}}}& \textbf{MRR} & \textbf{H@1} & \textbf{H@3} & \multicolumn{1}{c|}{\textbf{H@5}} & \textbf{MRR} & \textbf{H@1} & \textbf{H@3} & \multicolumn{1}{c|}{\textbf{H@5}} & \textbf{MRR}  & \textbf{H@1} & \textbf{H@3} & \textbf{H@5} \\
\hline
\multicolumn{1}{l|}{CLIP~\citep{Alec_2021}$^\lozenge$}   &88.23  &83.23  &92.10  &\multicolumn{1}{c|}{94.51}   &77.57 &67.78  &85.22  &\multicolumn{1}{c|}{90.04}  &
71.69  &61.21  &79.63  &85.18 \\ 
\multicolumn{1}{l|}{ViLT~\citep{Wonjae_2021}$^\lozenge$}   &79.46  &72.64  &84.51  &\multicolumn{1}{c|}{87.86}   &56.63 &45.85  &62.96  &\multicolumn{1}{c|}{69.80}  &45.22  &34.39  &51.07  &57.83 \\ 
\multicolumn{1}{l|}{ALBEF~\citep{Junnan_2021}$^\lozenge$}   &84.56  &78.64  &88.93  &\multicolumn{1}{c|}{91.75}   &75.29 &65.17  &82.84  &\multicolumn{1}{c|}{88.28}  &69.93  &60.59  &75.59  &81.30 \\ 
\multicolumn{1}{l|}{METER~\citep{Zi_2022}$^\lozenge$}   &79.49  &72.46  &84.41  &\multicolumn{1}{c|}{88.17}   &74.15 &63.96  &82.24  &\multicolumn{1}{c|}{87.08}  &63.71  &53.14  &70.93  &77.59 \\ 
\hline\hline
\multicolumn{1}{l|}{GPT-3.5-Turbo~\citep{Senbao_2024}$^\blacklozenge$}   &--  &73.80  &--  &\multicolumn{1}{c|}{--}   &-- &--  &--  &\multicolumn{1}{c|}{--}  &$\times$  &$\times$  &$\times$  &$\times$ \\ 
\multicolumn{1}{l|}{GEMEL~\citep{Senbao_2024}$^\blacklozenge$}   &--  &82.60  &--  &\multicolumn{1}{c|}{--}   &-- &--  &--  &\multicolumn{1}{c|}{--}  &$\times$  &$\times$  &$\times$  &$\times$ \\ 
\multicolumn{1}{l|}
{GELR~\citep{Xinwei_24}$^\blacklozenge$}   &--  &84.80  &--  &\multicolumn{1}{c|}{--}   &-- &--  &--  &\multicolumn{1}{c|}{--}  &$\times$  &$\times$  &$\times$  &$\times$ \\
\hline\hline
\multicolumn{1}{l|}{DZMNED~\citep{Seungwhan_2018}$^\lozenge$}   &84.97  &78.82  &90.02  &\multicolumn{1}{c|}{92.62}   &76.63 &68.16  &82.94  &\multicolumn{1}{c|}{87.33}  &67.59  &56.90  &75.34  &81.41 \\ 
\multicolumn{1}{l|}{JMEL~\citep{Omar_2020}$^\lozenge$}   &73.39  &64.65  &79.99  &\multicolumn{1}{c|}{84.34}   &60.06 &48.82  &66.77  &\multicolumn{1}{c|}{73.99}  & 48.19  &37.38  &54.23  &61.00 \\ 
\multicolumn{1}{l|}{VELML~\citep{Qiushuo_2022}$^\lozenge$}   &83.42  &76.62  &88.75  &\multicolumn{1}{c|}{91.96}   &77.19 &67.71  &84.57  &\multicolumn{1}{c|}{89.17}  & 66.13  &54.56  &74.43  &81.15 \\ 
\multicolumn{1}{l|}{GHMFC~\citep{Peng_2022}$^\lozenge$}   &83.36  &76.55  &88.40  &\multicolumn{1}{c|}{92.01}   &80.76 &72.92  &86.85  &\multicolumn{1}{c|}{90.60}  & 70.99  &60.27  &79.40  &84.74 \\
\multicolumn{1}{l|}{FissFuse~\citep{Pengfei_2024}$^\blacklozenge$}   &92.02  &87.89  &95.36  &\multicolumn{1}{c|}{--}   &-- &--  &--  &\multicolumn{1}{c|}{--}  &$\times$  &$\times$  &$\times$  &$\times$ \\
\multicolumn{1}{l|}{MELOV~\citep{Xuhui_2024}$^\blacklozenge$}   &92.32  &88.91  &95.61  &\multicolumn{1}{c|}{96.58}   &88.80 &84.14  &92.81  &\multicolumn{1}{c|}{94.89}  &76.57  &67.32  &83.69  &87.54 \\
\multicolumn{1}{l|}{OT-MEL~\citep{Zefeng_2024}$^\blacklozenge$}   &92.59  &88.97  &95.63  &\multicolumn{1}{c|}{96.96}   &88.27 &83.30  &92.39  &\multicolumn{1}{c|}{94.83}  &75.43  &66.07  &82.82  &87.39 \\
\hline
\multicolumn{1}{c|}{MIMIC~\citep{Pengfei_2023}$^\lozenge$}   &91.82  &87.98  &95.07  &\multicolumn{1}{c|}{96.37}   &86.95 &81.02  &91.77  &\multicolumn{1}{c|}{94.38}  &73.44  &63.51  &81.04  &86.43 \\
\multicolumn{1}{c|}
{MIMIC+DME}   &92.37  &88.66  &95.67  &\multicolumn{1}{c|}{97.10}   &87.49 &81.64  &92.22  &\multicolumn{1}{c|}{94.69}  &82.38  &75.70  &87.30  &90.57 \\
\multicolumn{1}{c|}{$\Delta$}   &\textcolor{mygreen}{+0.55}  &\textcolor{mygreen}{+0.68} &\textcolor{mygreen}{+0.60} &\multicolumn{1}{c|}{\textcolor{mygreen}{+0.73}} &\textcolor{mygreen}{+0.54} &\textcolor{mygreen}{+0.62} &\textcolor{mygreen}{+0.45} &\multicolumn{1}{c|}{\textcolor{mygreen}{+0.31}} &\textcolor{mygreen}{+8.94} &\textcolor{mygreen}{+12.19} &\textcolor{mygreen}{+6.26} &\textcolor{mygreen}{+4.14}\\
\hline
\multicolumn{1}{c|}{M${\rm^3}$EL~\citep{Zhiwei_2024}$^\blacklozenge$}   &92.30  &88.84  &95.20  &\multicolumn{1}{c|}{96.71}   &88.26 &82.82  &92.73  &\multicolumn{1}{c|}{\cellcolor{magenta!30}\textbf{95.34}}  &81.29  &74.06  &86.57  &90.04 \\
\multicolumn{1}{c|}{M${\rm^3}$EL+DME}   &92.51  &88.91  &95.49  &\multicolumn{1}{c|}{96.98}   &88.84 &83.89  &92.84  &\multicolumn{1}{c|}{94.95}  &83.53  &76.71  &88.79  &92.25 \\
\multicolumn{1}{c|}{$\Delta$}   &\textcolor{mygreen}{+0.21}  &\textcolor{mygreen}{+0.07} &\textcolor{mygreen}{+0.29} &\multicolumn{1}{c|}{\textcolor{mygreen}{+0.27}} &\textcolor{mygreen}{+0.58} &\textcolor{mygreen}{+1.07} &\textcolor{mygreen}{+0.11} &\multicolumn{1}{c|}{\textcolor{myred}{-0.39}} &\textcolor{mygreen}{+2.24} &\textcolor{mygreen}{+2.65} &\textcolor{mygreen}{+2.22} &\textcolor{mygreen}{+2.21}\\
\hline
\multicolumn{1}{c|}{MMoE}   &92.53  &89.07  &95.36  &\multicolumn{1}{c|}{97.02}   &89.04 &84.36  &92.93  &\multicolumn{1}{c|}{94.64}  &81.72  &74.59  &87.25  &90.33 \\
\multicolumn{1}{c|}{MMoE+DME}   &\cellcolor{magenta!30}\textbf{93.75} &\cellcolor{magenta!30}\textbf{90.77}  &\cellcolor{magenta!30}\textbf{96.27}  &\multicolumn{1}{c|}{\cellcolor{magenta!30}\textbf{97.41}}   &\cellcolor{magenta!30}\textbf{89.86} &\cellcolor{magenta!30}\textbf{85.51}  &\cellcolor{magenta!30}\textbf{93.43}  &\multicolumn{1}{c|}{95.31}  &\cellcolor{magenta!30}\textbf{84.23}  &\cellcolor{magenta!30}\textbf{77.57}  &\cellcolor{magenta!30}\textbf{89.12}  &\cellcolor{magenta!30}\textbf{92.49} \\
\multicolumn{1}{c|}{$\Delta$}   &\textcolor{mygreen}{+1.22}  &\textcolor{mygreen}{+1.70} &\textcolor{mygreen}{+0.91} &\multicolumn{1}{c|}{\textcolor{mygreen}{+0.39}} &\textcolor{mygreen}{+0.82} &\textcolor{mygreen}{+1.15} &\textcolor{mygreen}{+0.50} &\multicolumn{1}{c|}{\textcolor{mygreen}{+0.67}} &\textcolor{mygreen}{+2.51} &\textcolor{mygreen}{+2.98} &\textcolor{mygreen}{+1.87} &\textcolor{mygreen}{+2.16}\\
\hline
\multicolumn{1}{c|}{\textit{improvement} (\%)}   &\textcolor{mygreen}{+1.16}  &\textcolor{mygreen}{+1.70} &\textcolor{mygreen}{+0.60} &\multicolumn{1}{c|}{\textcolor{mygreen}{+0.31}} &\textcolor{mygreen}{+0.82} &\textcolor{mygreen}{+1.15} &\textcolor{mygreen}{+0.50} &\multicolumn{1}{c|}{\textcolor{myred}{-0.03}} &\textcolor{mygreen}{+0.70} &\textcolor{mygreen}{+0.86} &\textcolor{mygreen}{+0.33} &\textcolor{mygreen}{+0.24}\\
\hline
\end{tabular*}
\caption{{{Evaluation of different models on three MEL datasets, $^\lozenge$ results are from ~\citep{Pengfei_2023}, $^\blacklozenge$ results are from the corresponding papers. Best scores are highlighted in \colorbox{magenta!30}{\textbf{bold}}. $\Delta$ represents the performance improvement before and after adding the DME module. The \textit{improvement} (\%) represents the performance improvement over the state-of-the-art modes. -- means that the corresponding models does not provide the corresponding dataset results, and $\times$ means that the dataset size used by the corresponding models is different from other baselines and cannot be directly compared. FissFuse~\citep{Pengfei_2024}, MELOV~\citep{Xuhui_2024}, and OT-MEL~\citep{Zefeng_2024} do not have open source code, so we could not perform +DME experiments.}}}
\label{table_main_results}
\end{table*}

\smallskip
\noindent\textbf{$\triangleright$ Baselines and Evaluations.}
{We compare our method with three types of baselines. ($i$) \textit{vision-and-language pre-trained methods}: including \textbf{CLIP}~\citep{Alec_2021}, \textbf{ViLT}~\citep{Wonjae_2021}, \textbf{ALBEF}~\citep{Junnan_2021}, and \textbf{METER}~\citep{Zi_2022}; ($ii$) \textit{generative-based methods}: including \textbf{GPT-3.5}~\citep{OpenAI_2023}, \textbf{GEMEL}~\citep{Senbao_2024}, and {\textbf{GELR}~\citep{Xinwei_24}}; ($iii$) \textit{multimodal interaction based methods}: including \textbf{DZMNED}~\citep{Seungwhan_2018}, \textbf{JMEL}~\citep{Omar_2020}, \textbf{VELML}~\citep{Qiushuo_2022}, \textbf{GHMFC}~\cite{Peng_2022}, \textbf{MIMIC}~\citep{Pengfei_2023}, {\textbf{FissFuss}~\citep{Pengfei_2024}, \textbf{M${\rm^3}$EL}~\citep{Zhiwei_2024}, \textbf{MELOV}~\citep{Xuhui_2024}} and \textbf{OT-MEL}~\citep{Zefeng_2024}. {We note  that although DWE~\citep{Shezheng_2024} and UniMEL~\citep{Qi_2024} also tackle the MEL task, they use different dataset sizes than that from the general benchmark~\citep{Pengfei_2023, Qi_2024, Zhiwei_2024, Xuhui_2024, Zefeng_2024} and different evaluation indicators. Thus, we do not consider them as baselines.} Consistent with all baselines, we use the MRR and Hits@\textit{n} (abbreviated as H@\textit{n}, $n \in \{1, 3, 5\}$) as the evaluation metrics. Details about the description of baselines can be found in \textcolor{magenta}{\textbf{\hyperlink{baselines}{\textcolor{blue}{Appendix A}}}}. }

\subsection{Main Results}
\label{main_results}
{To address \textbf{RQ1}, we conduct experiments on the three datasets, the corresponding results are shown in Table~\ref{table_main_results} and Table~\ref{table_low_resource}.}

\smallskip
\noindent\textbf{$\triangleright$ Overall Performance.}
{Table~\ref{table_main_results} reports the performance of MMoE and baselines on three datasets. We can observe the following two findings: on the one hand, regardless of whether the DME module is incorporated, MMoE consistently achieves the best performance on all datasets, in particular, we observe that MMoE respectively achieves a 1.16\% and 1.70\% improvement on MRR and Hits@1 on WikiMEL compared to the best performing
baseline. On the other hand, incorporating the DME module into MIMIC, M${\rm^3}$EL and MMoE results in notable performance improvements, indicating the necessity of enriching semantic information within mention context. Especially for the WikiDiverse dataset, the model incorporating the DME module demonstrates a  significant improvement. We believe that the main reason is that in WikiDiverse the textual mention context not only has a shorter average sequence length (with an average of 11 words) but also contains a high number of abbreviations and acronyms. This greatly limits the availability of semantically relevant textual knowledge, leading to increased ambiguity when predicting the correct entities. DME module effectively incorporates WikiData entity descriptions into the semantic representation process of mentions, which complements the semantic knowledge of the mention context. Moreover, even without the DME module, MMoE still achieves the best performance under the same experimental conditions, and it attains optimal results when the DME is integrated with the MoE mechanism, underscoring the critical role of the interplay between DME and the other MMoE components.}

\smallskip
\noindent\textbf{$\triangleright$ Low Resource Settings.}
{Following~\citep{Pengfei_2023, Zhiwei_2024}, we conduct additional experiments on the RichpediaMEL and WikiDiverse datasets in low-resource scenarios, where only  10\% and 20\% of the training data is used as the new training set while keeping the validation and test sets unchanged. The corresponding experimental results are shown in Table~\ref{table_low_resource}. We have the following two main findings: on the one hand, MMoE consistently achieves the best performance, however, no baseline consistently ranks second, indicating distinct model focuses and overly tight model-dataset coupling. On the other hand, for the RichpediaMEL dataset, the performance gains obtained by increasing the data  from 10\% to 20\% are limited. This might be due to the more complete modal knowledge available in RichpediaMEL. For instance, in RichpediaMEL, 89.04\% of the mentions come with visual images, whereas in WikiDiverse only 44.37\%.
}

\subsection{Ablation Studies}
\label{ablation_study}
{To address \textbf{RQ2}, we conduct ablation experiments to verify the contribution of each component of MMoE, including: a) impact of different losses; b) impact of different modules. The corresponding results for three datasets are shown in Table~\ref{table_ablation_studies}.}

\smallskip
\noindent\textbf{$\triangleright$ Impact of Different Losses.}
{The overall loss ${\mathcal{L}}$ in Equation 8 comprises four components: ${\mathcal{L}}_T$, ${\mathcal{L}}_V$, ${\mathcal{L}}_C$, and ${\mathcal{L}}_O$. We validate the necessity of each component by individually removing them. The experimental results are presented in the upper half of Table~\ref{table_ablation_studies}. It is evident that, despite the presence of some outliers, the removal of each loss component generally leads to a performance degradation. An example of an outlier is observed in the WikiDiverse dataset, where the exclusion of the visual modality-related loss results in a slight improvement. This phenomenon can be attributed to two potential reasons: on the one hand, the availability of visual information in WikiData is sparse, with only 44.37\% of the mentions containing visual  content, which is considerably lower than the 85.65\% and 89.04\% in WikiMEL and RichpediaMEL, respectively. On the other hand, the visual information may not be closely related to the mentions, introducing  noise that could adversely affect performance. Despite these data-induced outliers, the overall trend indicates that removing any loss component results in performance degradation.}

\begin{table*}[!htp]
\setlength{\abovecaptionskip}{0.23cm}
\renewcommand\arraystretch{1.15}
\setlength{\tabcolsep}{0.57em}
\centering
\small
\begin{tabular*}{\linewidth}{@{}ccccccccccccccccc@{}}
\hline
\multicolumn{1}{c|}{\multirow{1}{*}{\textbf{Methods}}} & \multicolumn{4}{c|}{\textbf{RichpediaMEL(10\%)}} & \multicolumn{4}{c|}{\textbf{RichpediaMEL(20\%)}} & \multicolumn{4}{c|}{\textbf{WikiDiverse(10\%)}} & \multicolumn{4}{c}{\textbf{WikiDiverse(20\%)}}\\
\cline{1-5}\cline{6-9}\cline{10-13}\cline{11-17}
\multicolumn{1}{c|}{\multirow{1}{*}{\textbf{Metrics}}} & \textbf{MRR} & \textbf{H@1} & \textbf{H@3} & \multicolumn{1}{c|}{\textbf{H@5}} & \textbf{MRR} & \textbf{H@1} & \textbf{H@3} & \multicolumn{1}{c|}{\textbf{H@5}} & \textbf{MRR}  & \textbf{H@1} & \textbf{H@3} & \multicolumn{1}{c|}{\textbf{H@5}} &
\textbf{MRR}  & \textbf{H@1} & \textbf{H@3} & \textbf{H@5} \\
\hline
\multicolumn{1}{c|}{DZMNED}  &31.79 	&22.57 	&34.95 	&\multicolumn{1}{c|}{41.33}  &47.01 	&36.38 	&52.25  &\multicolumn{1}{c|}{58.28}   &19.99 	&11.45 	&22.52 	&\multicolumn{1}{c|}{29.50}  &40.97 	&28.73   &47.35    &56.69 \\
\multicolumn{1}{c|}{JMEL}   &25.01 	&16.70 	&27.68 	&\multicolumn{1}{c|}{33.63}   &39.38  &28.92	&43.35 	&\multicolumn{1}{c|}{50.59}  &28.26 	&19.97 	&32.19 	&\multicolumn{1}{c|}{37.58}  &39.05 	&29.26 	&44.23 	&49.90 \\
\multicolumn{1}{c|}{VELML}   &35.52 	&27.15 	&38.60 	&\multicolumn{1}{c|}{43.99}  &59.24 	&48.85 	&64.91 	&\multicolumn{1}{c|}{71.76}  &40.70 	&30.51 	&46.20 	&\multicolumn{1}{c|}{52.36}  &54.76 	&43.65 	&61.36 	&67.66 \\
\multicolumn{1}{c|}{GHMFC}   &76.69 	&68.00 	&83.38 	&\multicolumn{1}{c|}{87.73}  &80.42 	&72.57 	&86.69 	&\multicolumn{1}{c|}{90.15}  &59.56 	&48.08 	&66.31 	&\multicolumn{1}{c|}{74.25}  &63.46 	&51.73 	&71.85 	&78.54 \\
\multicolumn{1}{c|}{MIMIC}   &74.62 	&64.49 	&82.03 	&\multicolumn{1}{c|}{87.59}  &82.73 	&75.60 	&88.63 	&\multicolumn{1}{c|}{91.72}  &69.70 	&60.54 	&76.18 	&\multicolumn{1}{c|}{81.33}  &63.46 	&61.01 	&77.67 	&83.35 \\
\multicolumn{1}{c|}{M${\rm^3}$EL}   &78.68 	&69.54 	&85.77 	&\multicolumn{1}{c|}{89.70}  &82.04 	&74.20 	&88.32 	&\multicolumn{1}{c|}{91.89}  &74.61 	&66.36 	&80.70 	&\multicolumn{1}{c|}{84.31}  &77.90 	&70.36 	&83.16 	&87.63 \\
\hline
\multicolumn{1}{c|}{MMoE}   &\cellcolor{magenta!30}\textbf{86.08} 	&\cellcolor{magenta!30}\textbf{80.63} 	&\cellcolor{magenta!30}\textbf{90.54} 	&\multicolumn{1}{c|}{\cellcolor{magenta!30}\textbf{93.12}}  &\cellcolor{magenta!30}\textbf{86.51} 	&\cellcolor{magenta!30}\textbf{81.19} 	&\cellcolor{magenta!30}\textbf{90.82} 	&\multicolumn{1}{c|}{\cellcolor{magenta!30}\textbf{93.18}}  &\cellcolor{magenta!30}\textbf{78.04} 	&\cellcolor{magenta!30}\textbf{70.93} 	&\cellcolor{magenta!30}\textbf{82.96} 	&\multicolumn{1}{c|}{\cellcolor{magenta!30}\textbf{86.72}}  &\cellcolor{magenta!30}\textbf{81.25} 	&\cellcolor{magenta!30}\textbf{74.25} 	&\cellcolor{magenta!30}\textbf{86.62} 	&\cellcolor{magenta!30}\textbf{89.89} \\
\hline
\end{tabular*}
\caption{{Evaluation of different models in low resource settings on four MEL datasets, all the baselines results are from~\citep{Zhiwei_2024}. FissFuse~\citep{Pengfei_2024}, MELOV~\citep{Xuhui_2024}, and OT-MEL~\citep{Zefeng_2024} do not provide access to their code, so low resource experimental results could not be obtained. }
}
\label{table_low_resource}
\end{table*}

\begin{table*}[!htp]
\setlength{\abovecaptionskip}{0.18cm}
\renewcommand\arraystretch{1.37}
\setlength{\tabcolsep}{0.54em}
\centering
\small
\begin{tabular*}{\linewidth}{@{}cccccccccc@{}}
\hline
\multicolumn{1}{c|}{\multirow{1}{*}{\textbf{Methods}}} & \multicolumn{3}{c|}{\textbf{WikiMEL}} & \multicolumn{3}{c|}{\textbf{RichpediaMEL}} & \multicolumn{3}{c}{\textbf{WikiDiverse}}\\
\cline{1-4}\cline{5-7}\cline{8-10}
\multicolumn{1}{c|}{\multirow{1}{*}{\textbf{Metrics}}}& \textbf{MRR} & \textbf{Hits@1} & \multicolumn{1}{c|}{\multirow{1}{*}{\textbf{Hits@3}}} & \textbf{MRR} & \textbf{Hits@1} & \multicolumn{1}{c|}{\multirow{1}{*}{\textbf{Hits@3}}} & \textbf{MRR}  & \textbf{Hits@1} & \textbf{Hits@3} \\
\hline
\multicolumn{1}{c|}{\textit{w/o} ${\mathcal{L}}_T$} &92.29 \redbox{\fontsize{6pt}{8pt}\selectfont $\downarrow$1.46}  &88.95 \redbox{\fontsize{6pt}{8pt}\selectfont $\downarrow$1.82}  &\multicolumn{1}{c|}{95.03 \redbox{\fontsize{6pt}{8pt}\selectfont $\downarrow$1.24}} &87.38 \redbox{\fontsize{6pt}{8pt}\selectfont $\downarrow$2.48} &82.03 \redbox{\fontsize{6pt}{8pt}\selectfont $\downarrow$3.48}  &\multicolumn{1}{c|}{91.77 \redbox{\fontsize{6pt}{8pt}\selectfont $\downarrow$1.66}}  &82.66 \redbox{\fontsize{6pt}{8pt}\selectfont $\downarrow$1.57}   &75.75 \redbox{\fontsize{6pt}{8pt}\selectfont $\downarrow$1.82}  &87.73 \redbox{\fontsize{6pt}{8pt}\selectfont $\downarrow$1.39} \\
\multicolumn{1}{c|}{\textit{w/o} ${\mathcal{L}}_V$} &92.17 \redbox{\fontsize{6pt}{8pt}\selectfont $\downarrow$1.58}  &88.82 \redbox{\fontsize{6pt}{8pt}\selectfont $\downarrow$1.95}  &\multicolumn{1}{c|}{94.74 \redbox{\fontsize{6pt}{8pt}\selectfont $\downarrow$1.53}} &88.45 \redbox{\fontsize{6pt}{8pt}\selectfont $\downarrow$1.41} &83.60 \redbox{\fontsize{6pt}{8pt}\selectfont $\downarrow$1.91}  &\multicolumn{1}{c|}{92.36 \redbox{\fontsize{6pt}{8pt}\selectfont $\downarrow$1.07}}  &84.88 \greenbox{\fontsize{6pt}{8pt}\selectfont $\uparrow$0.65}   &78.78 \greenbox{\fontsize{6pt}{8pt}\selectfont $\uparrow$0.21}  &89.89 \greenbox{\fontsize{6pt}{8pt}\selectfont $\uparrow$0.77} \\
\multicolumn{1}{c|}{\textit{w/o} ${\mathcal{L}}_C$} &91.55 \redbox{\fontsize{6pt}{8pt}\selectfont $\downarrow$2.20}  &88.16 \redbox{\fontsize{6pt}{8pt}\selectfont $\downarrow$2.61}  &\multicolumn{1}{c|}{94.27 \redbox{\fontsize{6pt}{8pt}\selectfont $\downarrow$2.00}} &85.30 \redbox{\fontsize{6pt}{8pt}\selectfont $\downarrow$4.56} &80.71 \redbox{\fontsize{6pt}{8pt}\selectfont $\downarrow$4.80}  &\multicolumn{1}{c|}{88.66 \redbox{\fontsize{6pt}{8pt}\selectfont $\downarrow$4.77}}  &83.38 \redbox{\fontsize{6pt}{8pt}\selectfont $\downarrow$0.85}   &76.85 \redbox{\fontsize{6pt}{8pt}\selectfont $\downarrow$0.72}  &88.59 \redbox{\fontsize{6pt}{8pt}\selectfont $\downarrow$0.53} \\
\multicolumn{1}{c|}{\textit{w/o} ${\mathcal{L}}_O$} &92.53 \redbox{\fontsize{6pt}{8pt}\selectfont $\downarrow$1.22}  &89.07 \redbox{\fontsize{6pt}{8pt}\selectfont $\downarrow$1.70}  &\multicolumn{1}{c|}{95.36 \redbox{\fontsize{6pt}{8pt}\selectfont $\downarrow$0.91}} &89.04 \redbox{\fontsize{6pt}{8pt}\selectfont $\downarrow$0.82} &84.36 \redbox{\fontsize{6pt}{8pt}\selectfont $\downarrow$1.15}  &\multicolumn{1}{c|}{92.93 \redbox{\fontsize{6pt}{8pt}\selectfont $\downarrow$0.50}}  &83.66 \redbox{\fontsize{6pt}{8pt}\selectfont $\downarrow$0.57}   &77.09 \redbox{\fontsize{6pt}{8pt}\selectfont $\downarrow$0.48}  &88.88 \redbox{\fontsize{6pt}{8pt}\selectfont $\downarrow$0.24} \\
\hline
\multicolumn{1}{c|}{\textit{w/o} \texttt{IntraMoE}-\texttt{T}} &88.05 \redbox{\fontsize{6pt}{8pt}\selectfont $\downarrow$5.70}  &84.50 \redbox{\fontsize{6pt}{8pt}\selectfont $\downarrow$6.27}  &\multicolumn{1}{c|}{90.54 \redbox{\fontsize{6pt}{8pt}\selectfont $\downarrow$5.73}} &78.64 \redbox{\fontsize{6pt}{8pt}\selectfont $\downarrow$11.22} &73.67 \redbox{\fontsize{6pt}{8pt}\selectfont $\downarrow$11.84}  &\multicolumn{1}{c|}{81.47 \redbox{\fontsize{6pt}{8pt}\selectfont $\downarrow$11.96}}  &66.08 \redbox{\fontsize{6pt}{8pt}\selectfont $\downarrow$18.15}   &56.69 \redbox{\fontsize{6pt}{8pt}\selectfont $\downarrow$20.88}  &71.75 \redbox{\fontsize{6pt}{8pt}\selectfont $\downarrow$17.37} \\
\multicolumn{1}{c|}{\textit{w/o} \texttt{IntraMoE}-\texttt{V}} &90.36 \redbox{\fontsize{6pt}{8pt}\selectfont $\downarrow$3.39}  &86.32 \redbox{\fontsize{6pt}{8pt}\selectfont $\downarrow$4.45}  &\multicolumn{1}{c|}{93.31 \redbox{\fontsize{6pt}{8pt}\selectfont $\downarrow$2.96}} &87.64 \redbox{\fontsize{6pt}{8pt}\selectfont $\downarrow$2.22} &82.82 \redbox{\fontsize{6pt}{8pt}\selectfont $\downarrow$2.69}  &\multicolumn{1}{c|}{91.16 \redbox{\fontsize{6pt}{8pt}\selectfont $\downarrow$2.27}}  &84.29 \greenbox{\fontsize{6pt}{8pt}\selectfont $\uparrow$0.06}   &78.25 \greenbox{\fontsize{6pt}{8pt}\selectfont $\uparrow$0.68}  &88.88 \redbox{\fontsize{6pt}{8pt}\selectfont $\downarrow$0.24} \\
\multicolumn{1}{c|}{\textit{w/o} \texttt{InterMoE}} &91.41 \redbox{\fontsize{6pt}{8pt}\selectfont $\downarrow$2.34}  &87.95 \redbox{\fontsize{6pt}{8pt}\selectfont $\downarrow$2.82}  &\multicolumn{1}{c|}{94.00 \redbox{\fontsize{6pt}{8pt}\selectfont $\downarrow$2.27}} &88.30 \redbox{\fontsize{6pt}{8pt}\selectfont $\downarrow$1.56} &83.04 \redbox{\fontsize{6pt}{8pt}\selectfont $\downarrow$2.47}  &\multicolumn{1}{c|}{92.76 \redbox{\fontsize{6pt}{8pt}\selectfont $\downarrow$0.67}}  &82.66 \redbox{\fontsize{6pt}{8pt}\selectfont $\downarrow$1.57}   &75.99 \redbox{\fontsize{6pt}{8pt}\selectfont $\downarrow$1.58}  &87.58 \redbox{\fontsize{6pt}{8pt}\selectfont $\downarrow$1.54} \\
\hline
\multicolumn{1}{c|}{MMoE}   &\textbf{93.75}  &\textbf{90.77}  &\multicolumn{1}{c|}{\textbf{96.27}}   &\textbf{89.86} &\textbf{85.51}  &\multicolumn{1}{c|}{\textbf{93.43}}  &\textbf{84.23}  &\textbf{77.57}  &\textbf{89.12} \\
\hline
\end{tabular*}
\caption{{Evaluation of ablation studies on three MEL datasets.}}
\label{table_ablation_studies}
\end{table*}

\smallskip
\noindent\textbf{$\triangleright$ Impact of Different Modules.}
{To better understand the contribution to the performance of each module in the MMoE framework, we individually remove the \texttt{IntraMoE} for the textual and visual modalities, and the \texttt{InterMoE} for textual-visual modality interaction. The experimental results are presented in Table~\ref{table_ablation_studies}. The results show that removing module generally leads to performance degradation, with the most significant drop occurring when \texttt{IntraMoE-T} is removed. This is mainly because the textual information is more relevant for the MEL task than the visual one, thus having a greater impact on accuracy. Further analysis reveals that removing a module of a given  modality has a greater impact than only removing the corresponding modality's loss. {The primary reason lies in the fact that removing a modality implies that the corresponding modal knowledge is not fed into the model, whereas removing the modality's loss allows the knowledge to be passed into the model without explicit supervision on the modal knowledge. However, the modal knowledge still participates in the interactions among various modules of the model, implicitly contributing to the training process through these interactions.}}

\subsection{Parameter Sensitivity Analysis}
\begin{figure*}[htbp]
  \centering\subcaptionbox{Numbers of Experts and Top-experts}{
  \includegraphics[width=0.16\linewidth]{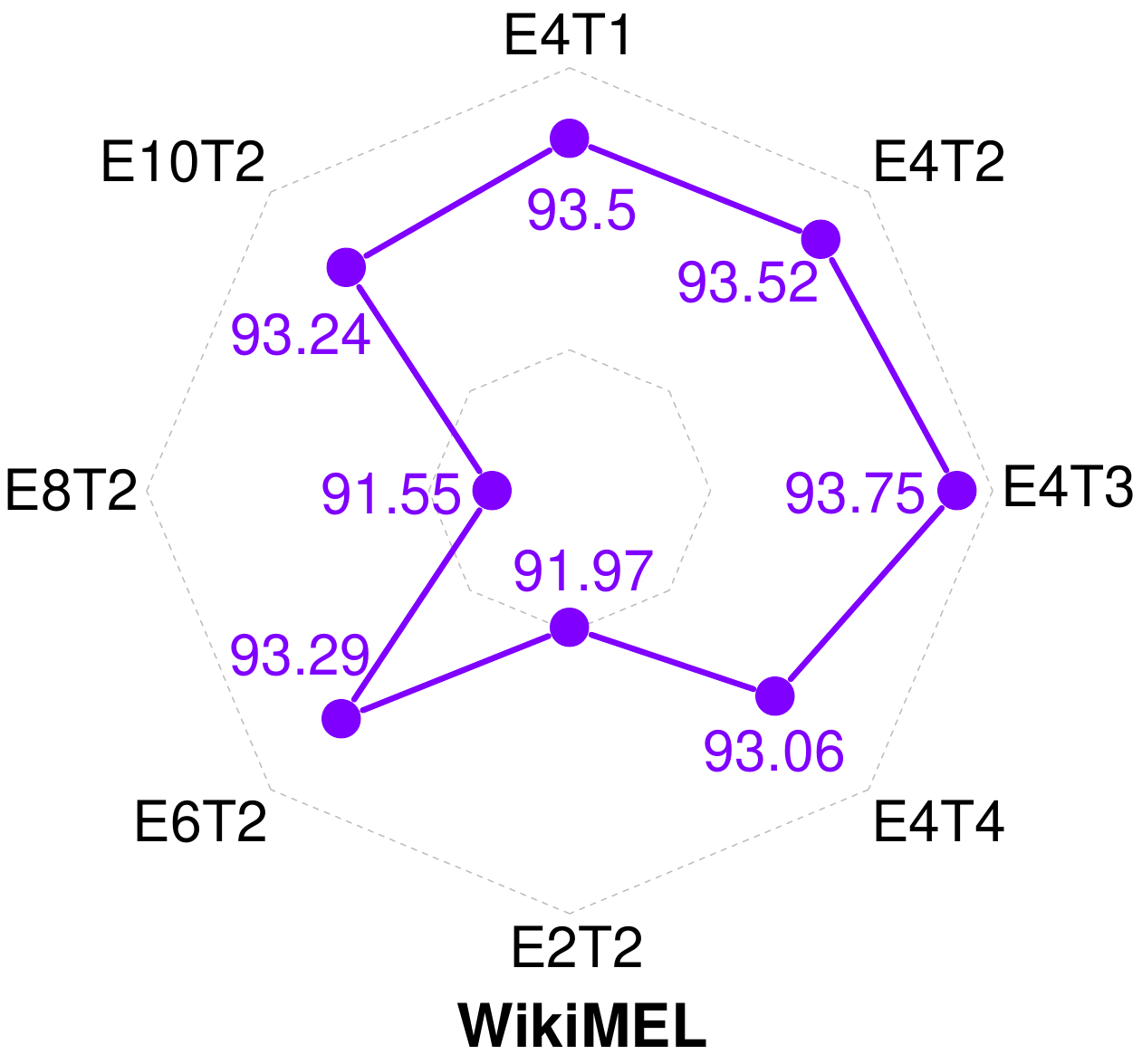}
  \includegraphics[width=0.16\linewidth]{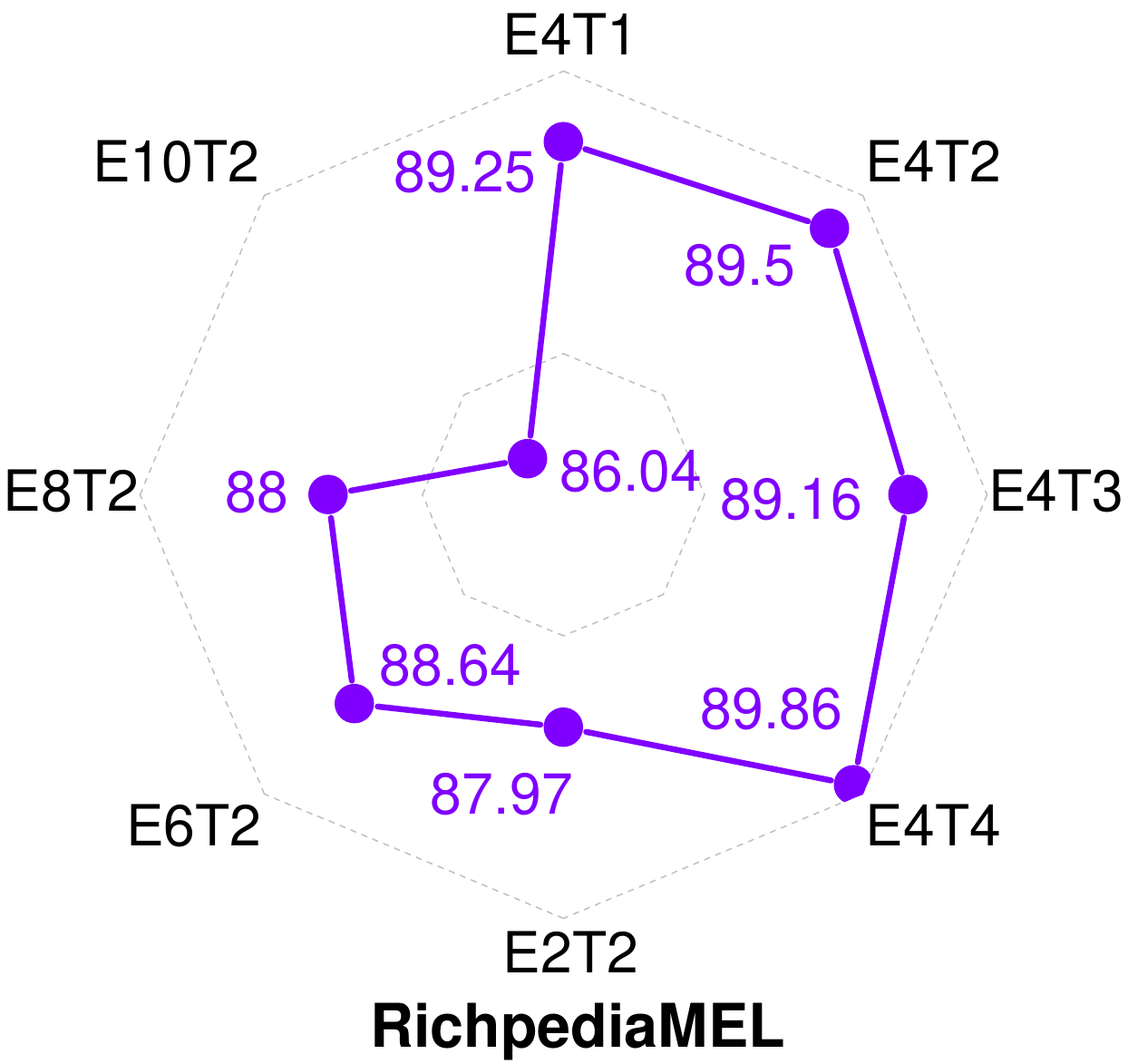}
  \includegraphics[width=0.16\linewidth]{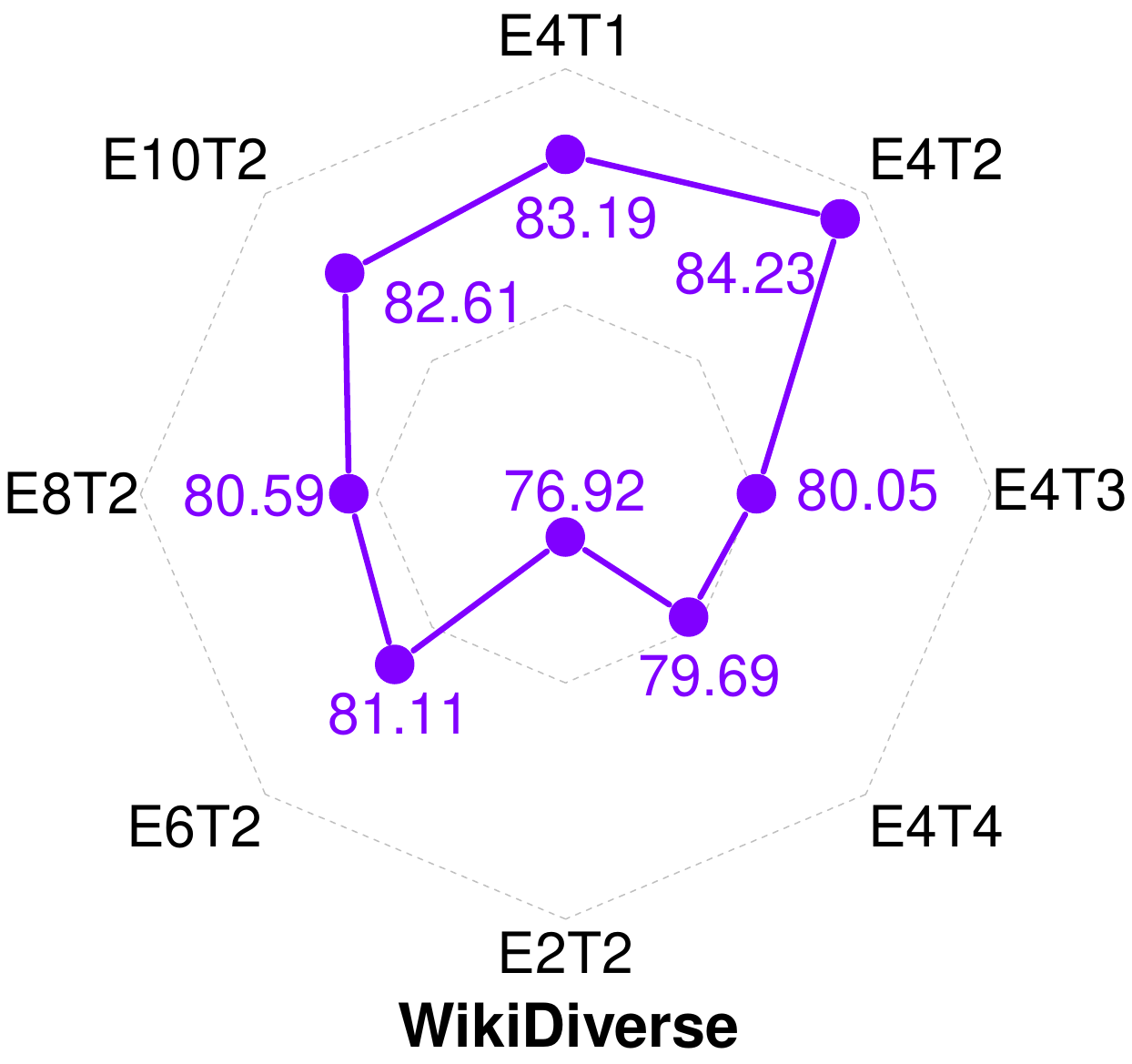}}\hfill
  \centering\subcaptionbox{Learning Rates}{
  \includegraphics[width=0.16\linewidth]{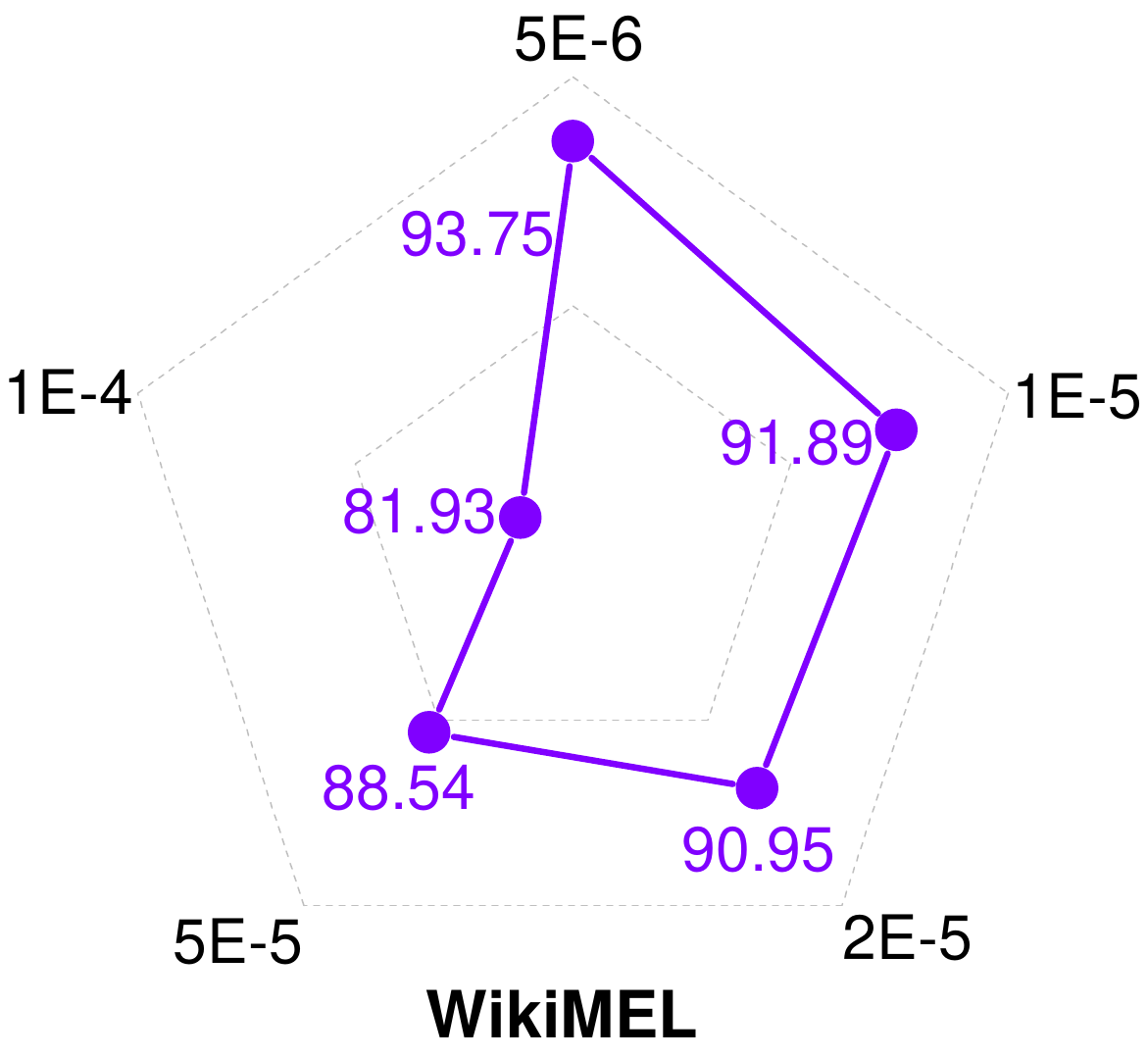}
  \includegraphics[width=0.16\linewidth]{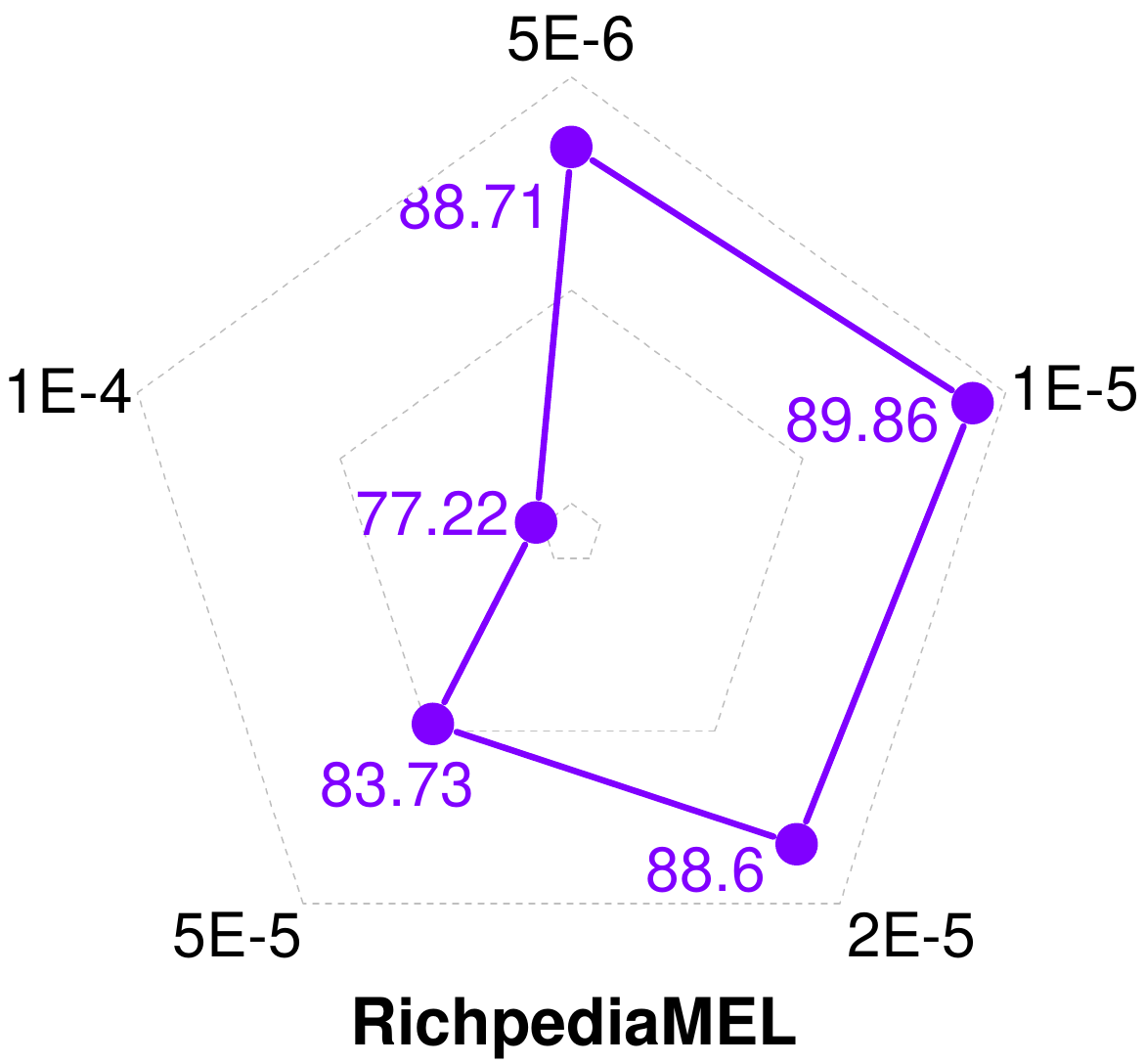}
  \includegraphics[width=0.16\linewidth]{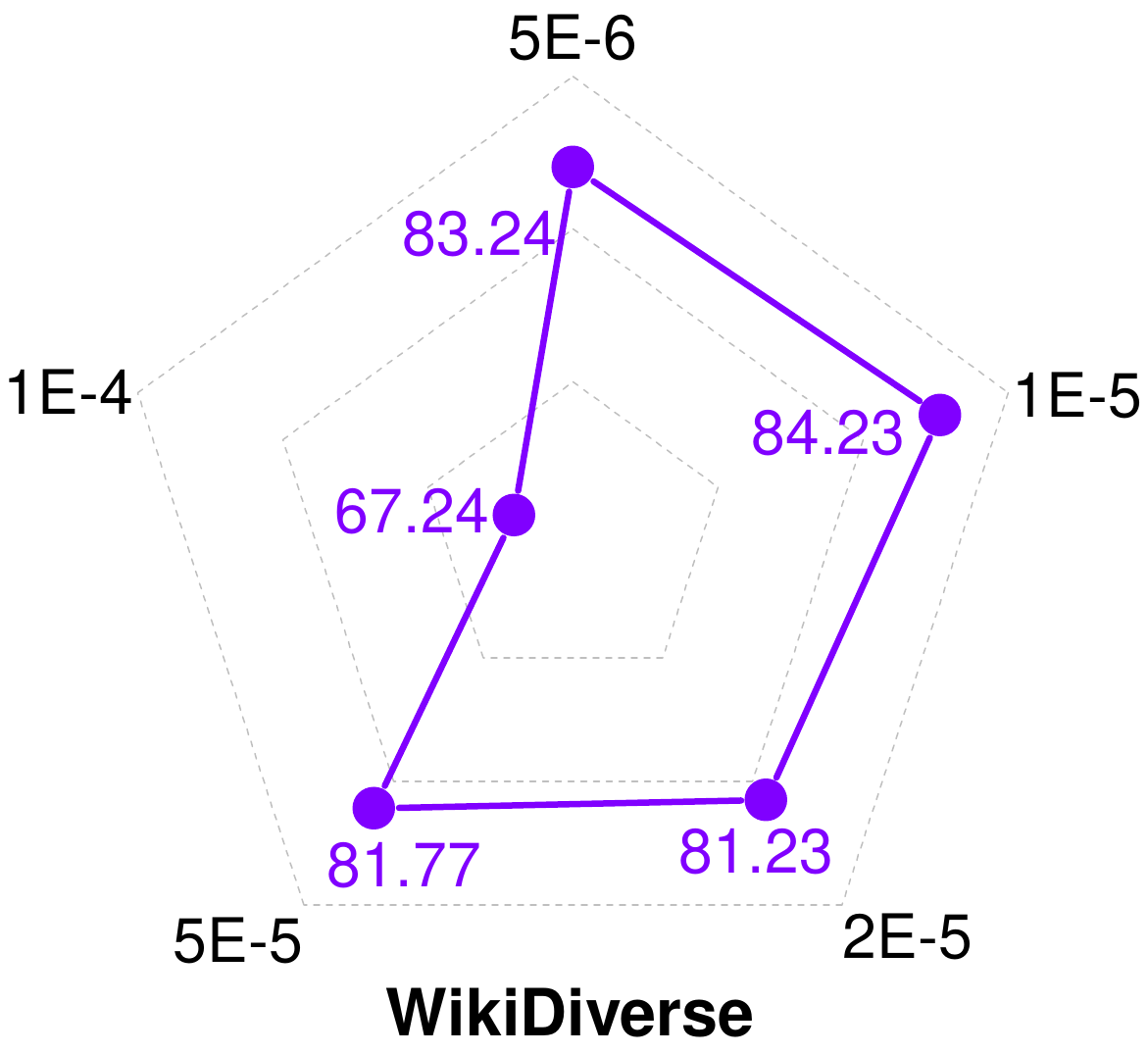}}\hfill
  \centering\subcaptionbox{Embedding Dimension}{
  \includegraphics[width=0.16\linewidth]{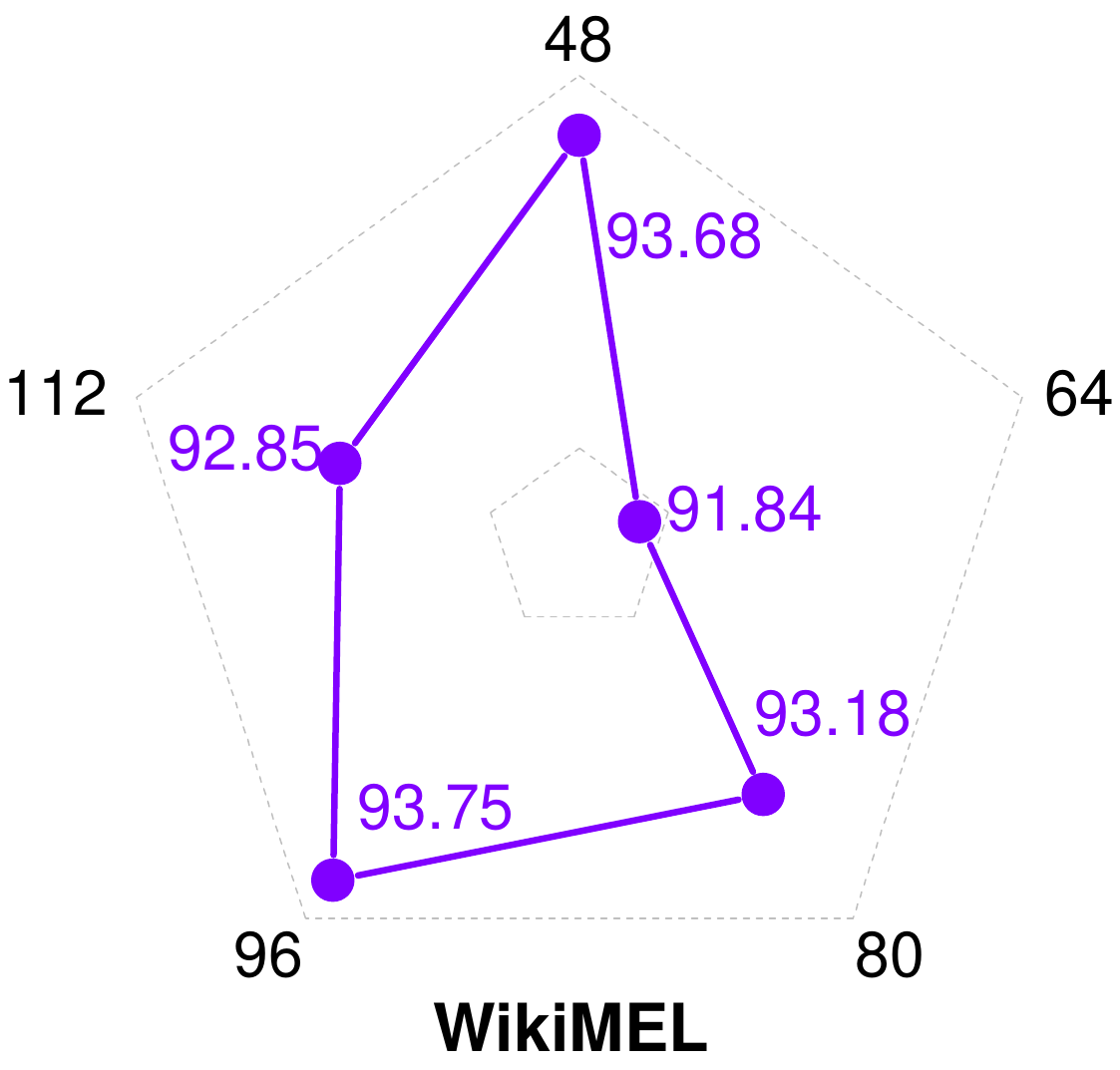}
  \includegraphics[width=0.16\linewidth]{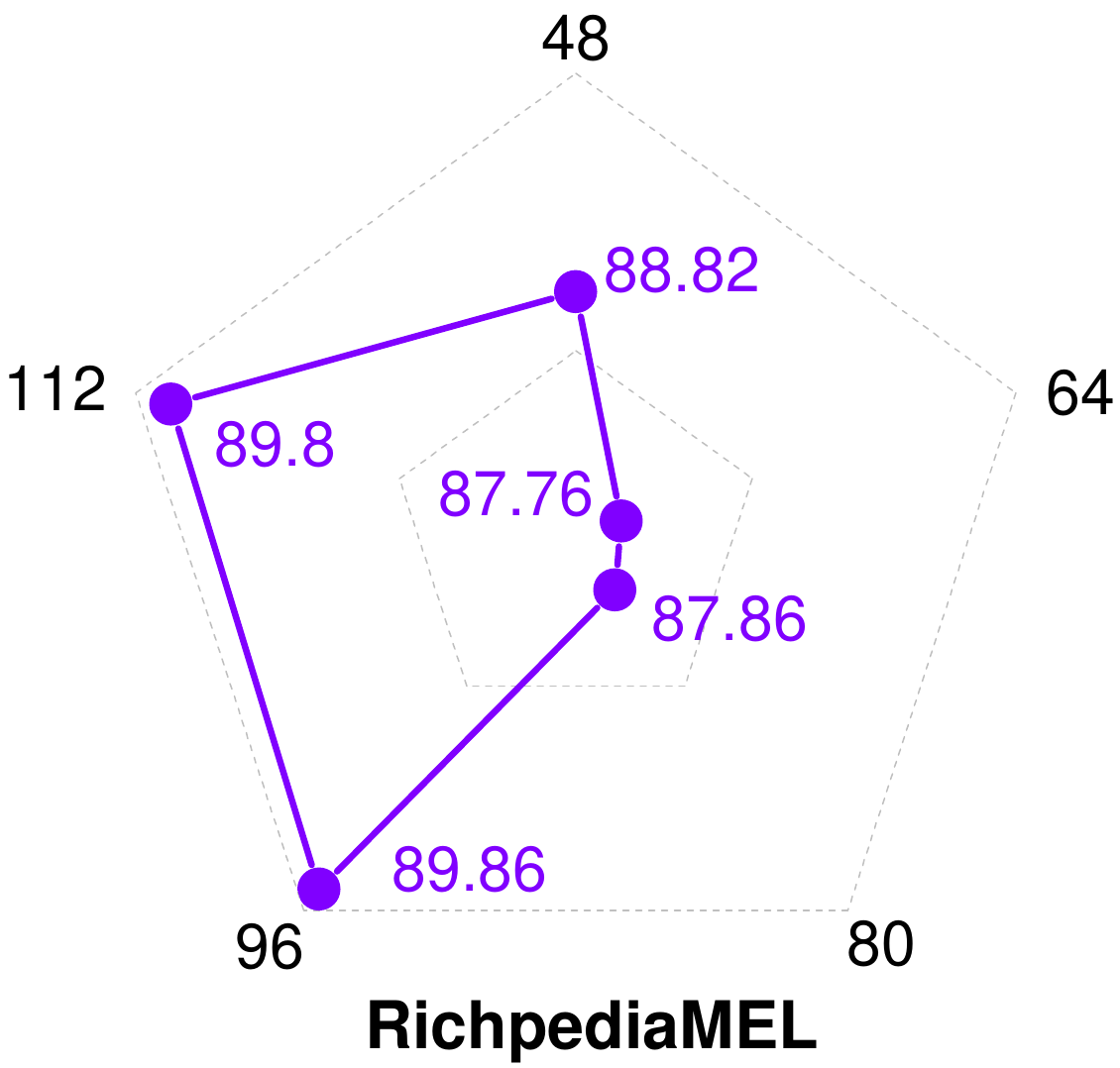}
  \includegraphics[width=0.16\linewidth]{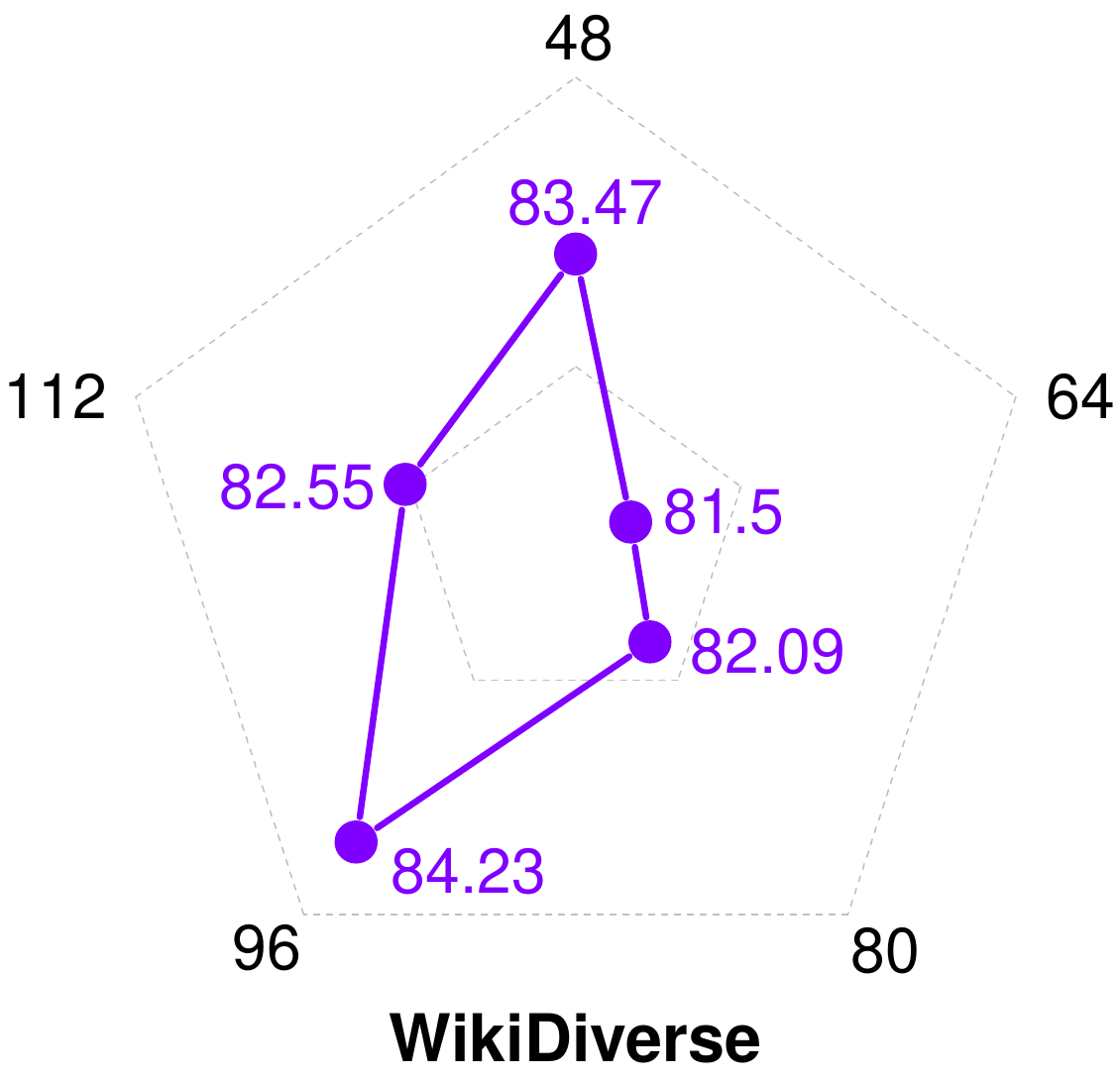}}\hfill
    \centering\subcaptionbox{Max Text Length}{
  \includegraphics[width=0.16\linewidth]{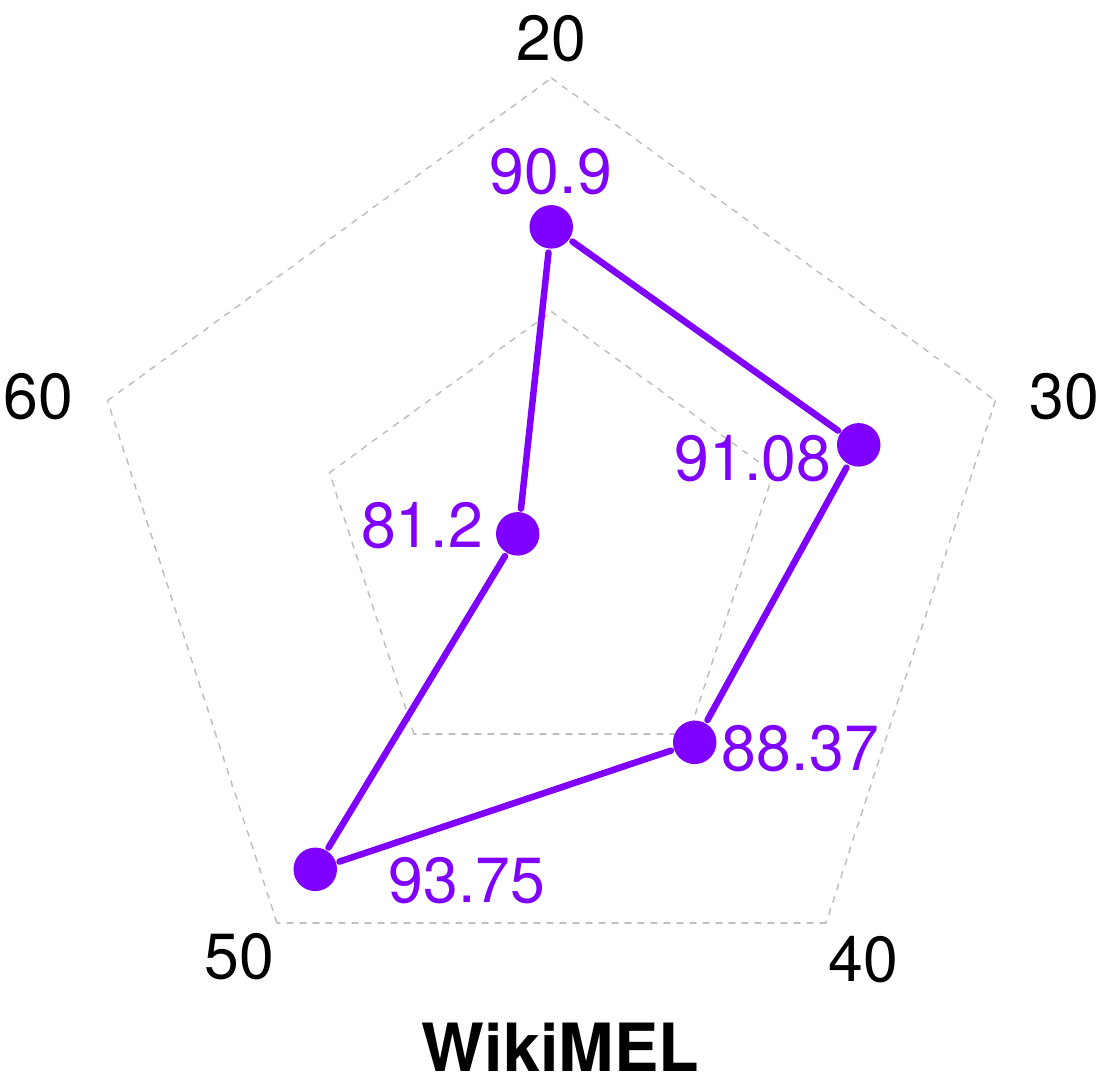}
  \includegraphics[width=0.16\linewidth]{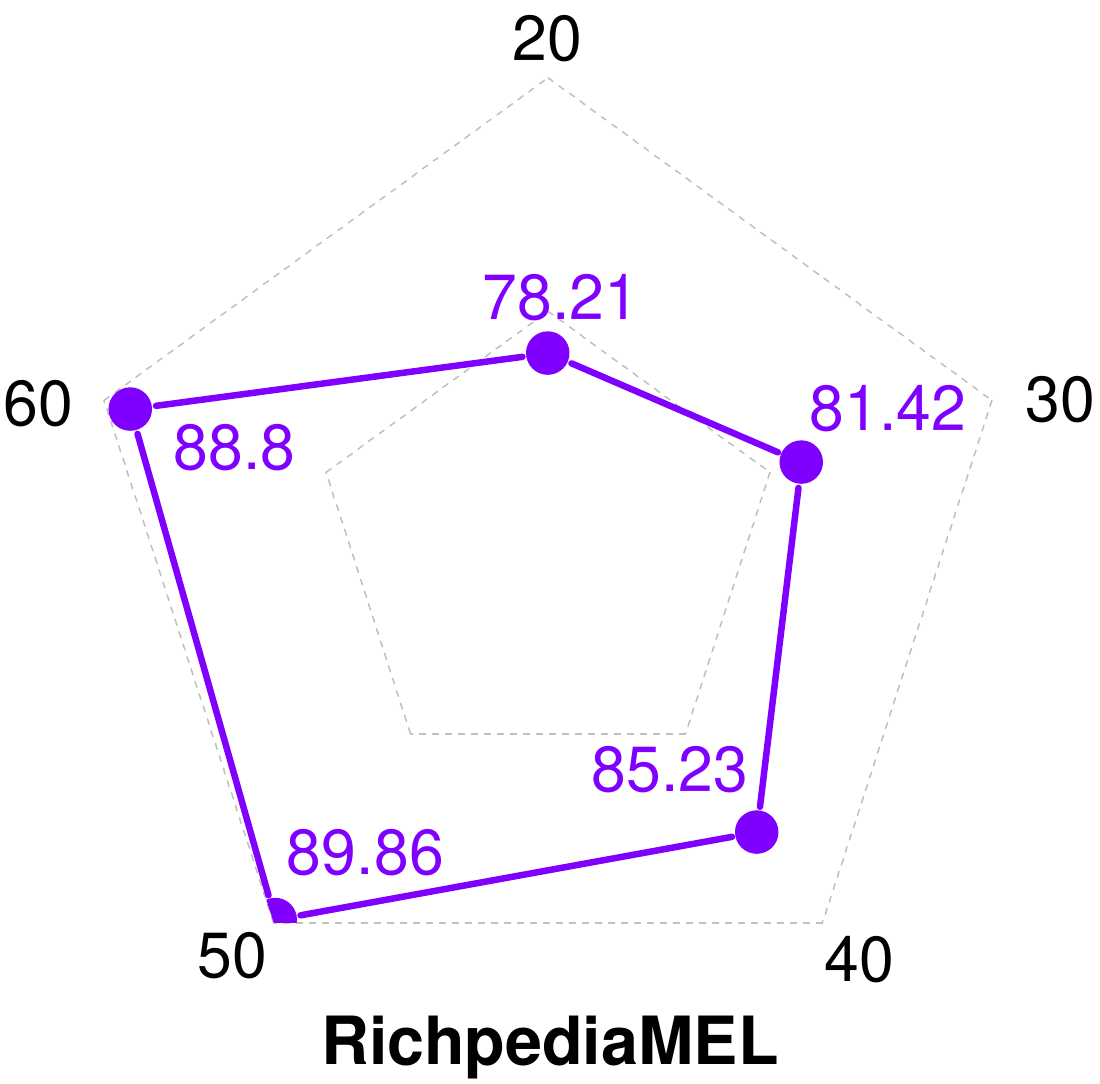}
  \includegraphics[width=0.16\linewidth]{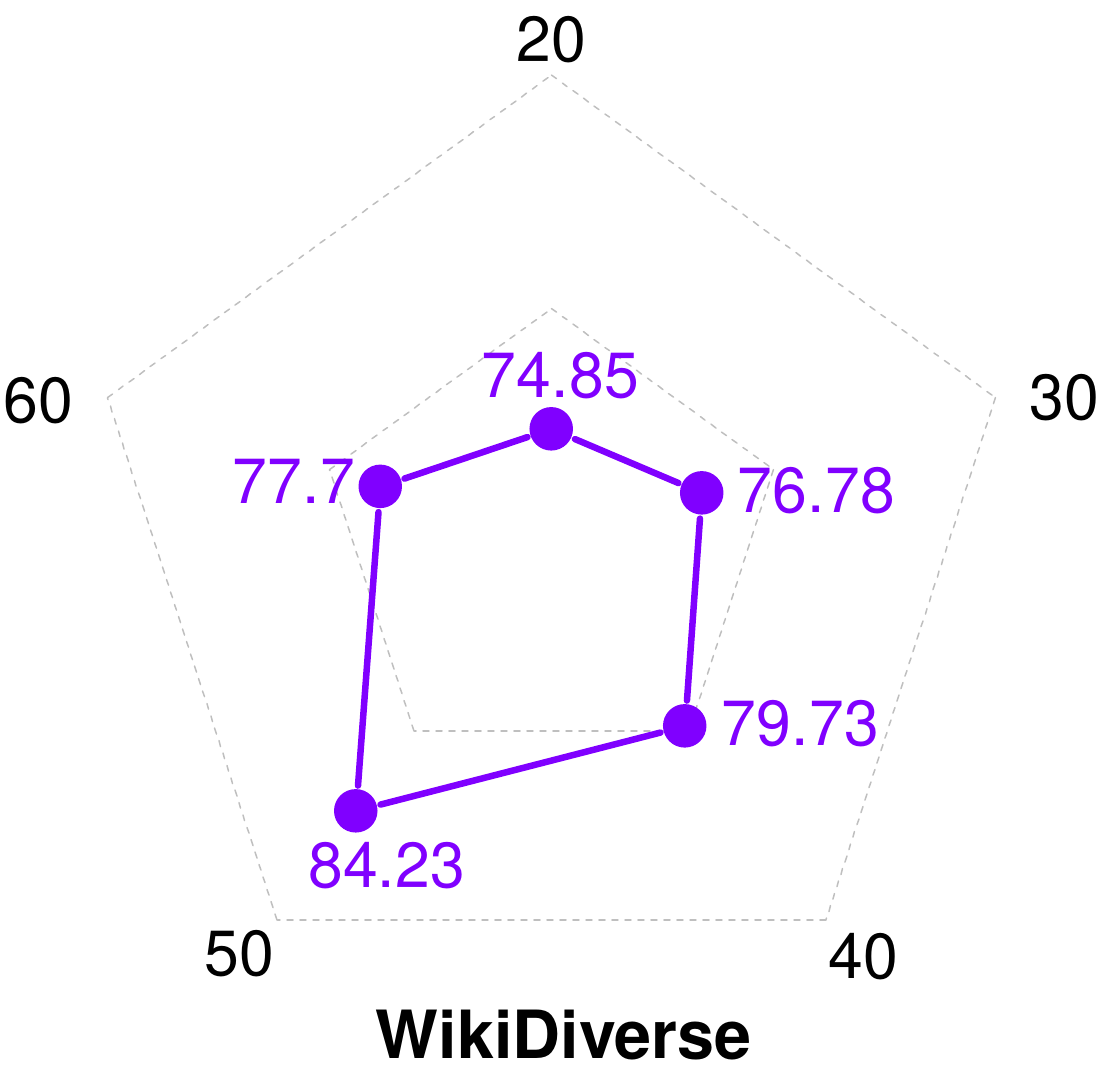}}\hfill
  \caption{Parameter sensitivity experiments under different conditions on WikiMEL, RichpediaMEL and WikiDiverse datasets.}
\label{figure_hyper_parameters}
\end{figure*}

{To address \textbf{RQ3}, we carry out parameter sensitivity experiments on the WikiMEL, RichpediaMEL and WikiDiverse datasets, mainly includes the following four aspects: a) effect of the number of experts and top-experts; b) effect of learning rates; c) effect of embedding dimension; and d) effect of max text length. The corresponding results for three datasets are shown in Figure~\ref{figure_hyper_parameters}.}

\smallskip
\noindent\textbf{$\triangleright$ Effect of the Numbers of Experts and Top-experts.}
\changed{In the SMoE mechanism, two critical hyperparameters, \textit{i.e.,} the number of experts and the number of top-experts jointly determine the model’s effectiveness and efficiency. The number of experts primarily affects the model's capacity and diversity, while the number of top-experts primarily influences the computational efficiency and the balance between performance and resource consumption. Selecting an appropriate number of experts and top-experts is crucial for balancing the model's performance and efficiency. We set the number of experts to \{$2$, $4$, $6$, $8$, $10$\} and the number of top-experts to \{$1$, $2$, $3$, $4$\}, creating $8$ experimental conditions to verify the impact of these parameters. The corresponding results on three datasets are shown in Figure~\ref{figure_hyper_parameters}(a). We can draw the following three conclusions: ($i$): Increasing the number of experts can enhance the model's expressive capacity to capture diverse features from different perspectives, but more experts are not always better. For instance, on the WikiMEL dataset, E$6$T$2$ outperforms E$8$T$2$; ($ii$): Top experts dynamically select suitable experts to process different samples by activating certain experts. The appropriate choice of top value is a trade-off between performance and resource efficiency. Generally, when memory capacity is large, we prefer to set a higher number of experts, but the number of top-experts is usually kept relatively small; ($iii$): Under the optimal conditions across the three datasets, the values for the number of experts and top experts differ. For WikiMEL, E$4$T$3$ achieves the best results, while for RichpediaMEL and WikiDiverse, E$4$T$4$ and E$4$T$2$ are optimal, respectively.}

\smallskip
\noindent\textbf{$\triangleright$ Effect of Learning Rates.}
\changed{A larger learning rate can expedite the convergence of the model but may lead to the omission of the optimal solution during the optimization process. Conversely, a lower learning rate facilitates a more stable convergence to the optimal solution, albeit at a slower rate. Therefore, selecting the learning rate involves a trade-off between convergence speed and optimal performance. We conducted experiments with five different learning rates across three datasets, with the corresponding results illustrated in Figure~\ref{figure_hyper_parameters}(b). We observed that the trends across the three datasets were generally consistent with respect to the learning rates: a learning rate of $1$E-$4$ resulted in the poorest performance, while a learning rate of $1$E-$5$ achieved the best performance. We attribute this to the stability of the chosen optimization algorithm.}

\smallskip
\noindent\textbf{$\triangleright$ Effect of Embedding Dimension.}
\changed{High-dimensional embeddings can capture more complex semantic information, enhancing the representational capacity of the model, but they also require more computational resources. In contrast, low-dimensional embeddings can effectively reduce the complexity of the model, which helps to prevent overfitting. We conduct a progressive comparison experiment on three datasets, selecting $5$ different dimensionality values. The corresponding results are shown in Figure~\ref{figure_hyper_parameters}(c). We can observe that when the embedding dimension is set to $96$, the model consistently achieves the best performance across all three datasets. Additionally, increasing the embedding dimension does not lead to an improvement in performance, which is consistent with the findings of~\citep{Xiao_2022}. In practice, it is crucial to select the smallest embedding dimension that yields optimal performance.}

\smallskip
\noindent\textbf{$\triangleright$ Effect of Max Text Length.}
\changed{The max length of text sequences in a mention context is a significant factor influencing the representational capacity of the textual modality. Longer texts contain a greater number of words, but they may also introduce more noisy-words that are less relevant to the mentions. Selecting an appropriate maximum text length requires a comprehensive consideration of the characteristics of the dataset itself, specifically analyzing the sequence lengths of the majority of data points in the dataset. We conducted comparative experiments with $5$ different maximum sequence lengths across three datasets, as shown in Figure~\ref{figure_hyper_parameters}(d). The results indicate that the performance improves with the increase of the length for all three datasets, peaking when the sequence length is set to $50$. Further increases in sequence length beyond this point, however, lead to a decline in performance. This is reasonable, as excessively short text sequences provide limited semantic coverage and insufficient support for predicting the corresponding entities, while excessively long sequences may disrupt the representation of mentions, leading to adverse effects.}

\subsection{Complexity Analysis}
We conduct a comparative analysis of the computational complexity of each module within the MMoE model, with the results detailed in Table~\ref{table_flops_and_params}. The computational complexity is assessed from two dimensions, the number of floating-point operations (\#FLOPs) and the number of parameters (\#Params). We observe that the SMoE module bears the majority of the computational burden, as evidenced by the significant reduction in complexity values upon its removal. This observation is reasonable, given that the SMoE mechanism involves a substantial number of feed-forward neural networks (FFNs), which are the primary contributors to both computational and temporal overhead. Furthermore, upon removing the IntraMoE and InterMoE modules, the computational costs were ranked as follows: \texttt{IntraMoE-T} $>$ \texttt{InterMoE} $>$ \texttt{IntraMoE-V}, with \texttt{IntraMoE-T} incurring in significantly higher costs than the other two. This is mainly due to the higher embedding dimensions in the intermediate layers of the SMoE module within \texttt{IntraMoE-T}. 

Additionally, we compared the training convergence times between the MMoE and MIMIC. On the WikiMEL, RichpediaMEL, and WikiDiverse datasets, the convergence times for MMoE are 1.88 hours, 1.53 hours, and 1.10 hours, respectively, while for the MIMIC, are 1.63 hours, 1.15 hours, and 0.95 hours, respectively. Although the MMoE requires relatively more time to train, this computational cost remains within an acceptable range.

\begin{table}[!htp]
\setlength{\abovecaptionskip}{0.03cm}
\renewcommand\arraystretch{1.2}
\setlength{\tabcolsep}{0.7em}
\centering
\small
\begin{tabular*}{\linewidth}{@{}cccc@{}}
\hline
\multicolumn{1}{c}{\textbf{\#FLOPs}}&\multicolumn{1}{c}{\textbf{WikiMEL}}&\multicolumn{1}{c}{\textbf{RichpediaMEL}}&\multicolumn{1}{c}{\textbf{WikiDiverse}} \\
\hline
\textit{w/o} \texttt{IntraMoE}-\texttt{T} &$3.135$G &$3.135$G &$3.135$G \\
\textit{w/o} \texttt{IntraMoE}-\texttt{V} &$18.435$G &$18.435$G &$18.435$G \\
\textit{w/o} \texttt{InterMoE} &$17.325$G &$17.325$G &$17.325$G \\
\textit{w/o} \texttt{SMoE} &$481.333$M &$481.333$M &$481.333$M \\
\texttt{MMoE} &$19.443$G &$19.443$G &$19.443$G \\
\hline
\multicolumn{1}{c}{\textbf{\#Params}}&\multicolumn{1}{c}{\textbf{WikiMEL}}&\multicolumn{1}{c}{\textbf{RichpediaMEL}}&\multicolumn{1}{c}{\textbf{WikiDiverse}} \\
\hline
\textit{w/o} \texttt{IntraMoE}-\texttt{T} &$683.849$K &$683.849$K &$683.849$K \\
\textit{w/o} \texttt{IntraMoE}-\texttt{V} &$5.378$M &$5.378$M &$5.378$M \\
\textit{w/o} \texttt{InterMoE} &$5.347$M &$5.347$M &$5.347$M \\
\textit{w/o} \texttt{SMoE} &$295.905$K &$295.905$K &$295.905$K \\
\texttt{MMoE} &$5.703$M &$5.703$M &$5.703$M \\
\hline
\end{tabular*}
\caption{Amount of parameters and calculation for MMoE model without various modules.}
\label{table_flops_and_params}
\end{table}

%% file: sections/conclusion.tex

{We introduced \texttt{MMoE} which simultaneously considers mention ambiguity and dynamic selection of different regions of information. We leverage LLMs to rank WikiData entity descriptions related to mentions to alleviate the issue of insufficient semantically relevant knowledge in mention contexts. 
In addition, we introduce a switch mixture-of-experts mechanism to dynamically specify the importance of different textual tokens and visual patches. Extensive empirical experiments and ablation studies
confirm MMoE's outstanding performance. 
In the future, we plan to explore tasks in specific domains, such as the biomedical and drug discovery domains. Moreover, the visual knowledge of three datasets is very sparse, thus how to complete it is an interesting research direction.}

\section*{Acknowledgments}
This work was supported by the National Natural Science Foundation of China (No.62376144), by the Science and Technology Cooperation and Exchange Special Project of Shanxi Province (No.202204041101016), by the Key Research and Development Project of Shanxi Province (No.202102020101008), by the Shanxi Natural Language Processing Innovation Team (Shanxi Talents) Project.

%% file: sections/appendix.tex
\subsection*{A\,\,\,Baselines}
\hypertarget{baselines}{}
\changed{We compare MMoE with three types of baselines, including \textit{vision-and-language pre-trained methods}, \textit{generative-based methods} and \textit{multimodal interaction based-methods}.}


The \textbf{\changed{vision-and-language pre-trained methods}} include:
\begin{itemize}[itemsep=0.5ex, leftmargin=5mm]
\item \textbf{CLIP}~\citep{Alec_2021} adopts BERT and ViT as the textual and visual embedding encoders, and designs a contrastive learning mechanism to bring paired text-image together, and pushes unpaired texts and images away from each other.
\item \textbf{ViLT}~\citep{Wonjae_2021} focuses on correlations between multiple modalities through a series of Transformer~\citep{Ashish_2017} layers.
\item \textbf{ALBEF}~\citep{Junnan_2021} uses a text-image contrastive loss to align textual and visual features and further combines them together using a multimodal Transformer encoder.
\item \textbf{METER}~\citep{Zi_2022} investigates how to train a performant vision-and-language Transformer, and  attributes the factors affecting model performance to a vision encoder, text encoder, multimodal fusion
module, and architectural design.
\end{itemize}


The \textbf{\changed{generative-based methods}} include:
\begin{itemize}[itemsep=0.5ex, leftmargin=5mm]
\item \textbf{GPT-3.5}~\citep{OpenAI_2023} is a powerful large language model developed by OpenAI.
\item \textbf{GEMEL}~\citep{Senbao_2024} keeps the vision and language model frozen and trains a feature mapper to enable cross-modality interactions, further applies a constrained decoding strategy to efficiently search the valid entity space.
\item {\textbf{GELR}~\citep{Xinwei_24} incorporates a knowledge retriever to enhance visual entity information by leveraging external sources, and devises a prioritization scheme to manage conflicts arising from the integration of external and internal knowledge.}
\end{itemize}

The \textbf{\changed{multimodal interaction based methods}} include:
\begin{itemize}[itemsep=0.5ex, leftmargin=5mm]
\item \textbf{DZMNED}~\citep{Seungwhan_2018} incorporates an attention mechanism to integrate textual-level, visual-level, and character-level features.
\item \textbf{JMEL}~\citep{Omar_2020} utilizes both unigram and bigram embeddings as textual features, and further concatenate these features to pass through a fully connected layer.
\item \textbf{VELML}~\citep{Qiushuo_2022} extracts visual features via  a VGG16~\citep{Karen_2014} network and replaces a GRU~\citep{Kyunghyun_2014} textual encoder with pre-trained BERT, and further designs a deep modal-attention mechanism to aggregate all modalities features.
\item \textbf{GHMFC}~\cite{Peng_2022} obtains the hierarchical features of textual and visual co-attention through the multi-modal co-attention mechanism, and proposes a contrastive loss to learn more generic multimodal features and reduce noise.
\item {\textbf{FissFuss}~\citep{Pengfei_2024} integrates the Fission and Fusion branches, establishing dynamic features for each mention-entity pair and adaptively learning multimodal interactions to alleviate content discrepancy.}
\item {\textbf{MELOV}~\citep{Xuhui_2024} incorporates inter-modality generation and intra-modality aggregation, effectively leveraging the shared information from heterogeneous textual features and relevant visual details of semantic similar neighbors.}
\item \textbf{OT-MEL}~\citep{Zefeng_2024} formulates the correlation assignment problem as an optimal transport problem, and exploits the multimodal correlation between mentions and entities to enhance the fine-grained matching.
\item \textbf{MIMIC}~\citep{Pengfei_2023} devises three different interaction matching modules to explore both intra-modal and inter-modal interactions among the features of entities and mentions.
\item {\textbf{M${\rm^3}$EL}~\citep{Zhiwei_2024} introduces an intra-modal contrastive learning mechanism to obtain better discriminative embedding representations, and devises intra-modal and cross-modal matching networks to explore different multimodal interactions.}
\end{itemize}

\begin{figure*}[!htp]
    \centering
    \includegraphics[width=1\textwidth]{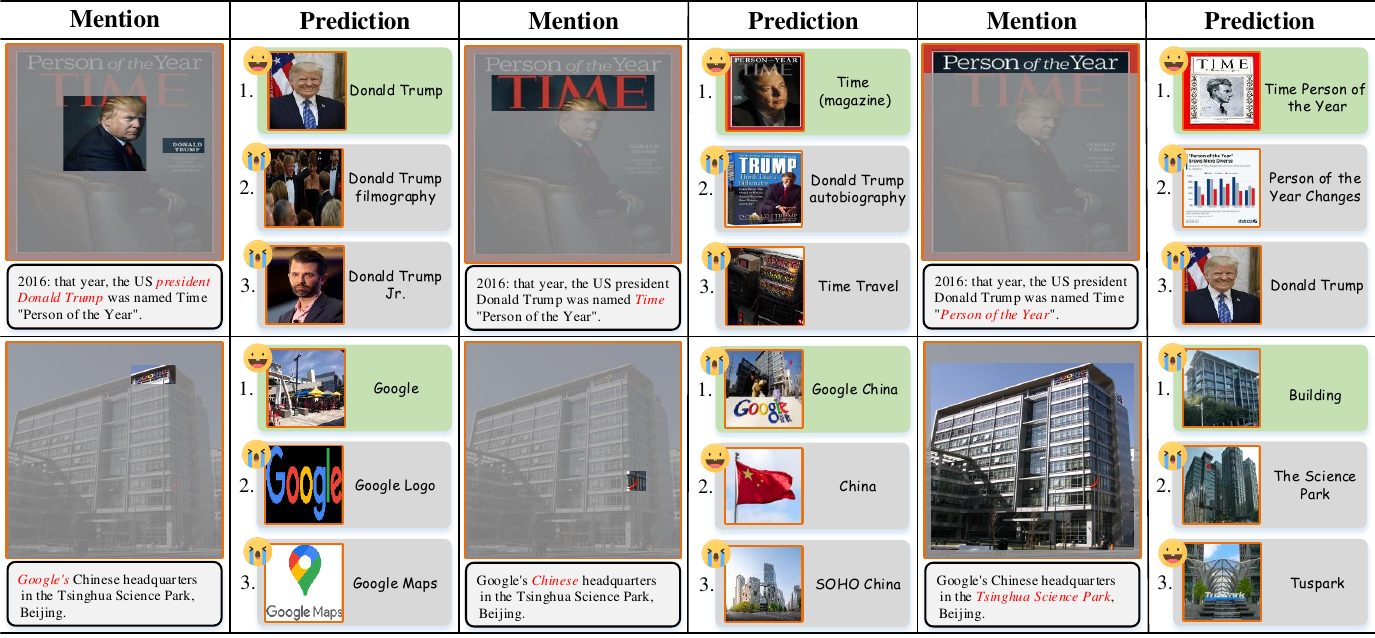}
    \caption{{Case study of  MMoE. These are two cases from the test set of WikiDiverse, each of them needing to predict three mentions with multimodal context. We display the top 3 ranked entities by  MMoE. To better showcase the result, we omit the descriptions of mentions.}}
    \label{figure_case_study}
\end{figure*}


\subsection*{B\,\,\,Case Study}
\hypertarget{case_study}{}
{To further quantitatively analyze the MMoE model, we select two test cases from the WikiDiverse dataset, with the top-3 predicted entities for each mention shown in Figure~\ref{figure_case_study}. Notably, both test cases require simultaneously predicting the entities corresponding to multiple mentions, yet they share identical visual images. This poses a challenge for existing MEL models to link mentions to the appropriate entities, as it requires filtering the most relevant visual knowledge based on the mentions to be predicted. However, MMoE effectively addresses the issue of fine-grained modal knowledge selection for mentions by dynamically selecting different patch regions of the visual modality according to the specific mentions being predicted. Furthermore, we observe that MMoE does not perform well for the prediction of mention \textit{Tsinghua Science Park}. The main reason is that the image  in this case depicts a building, which lacks any content relevant to mention \textit{Tsinghua Science Park}. Consequently, MMoE struggles to capture visual knowledge associated with \textit{Tsinghua Science Park}, making mispredictions inevitable.}

\begin{table*}[!htp]
\setlength{\abovecaptionskip}{0.18cm}
\renewcommand\arraystretch{1.0}
\setlength{\tabcolsep}{0.64em}
\centering
\small
\begin{tabular*}{\linewidth}{@{}ccccccccccccc@{}}
\hline
\multicolumn{1}{c|}{\multirow{1}{*}{\textbf{Methods}}} & \multicolumn{4}{c|}{\textbf{WikiMEL}} & \multicolumn{4}{c|}{\textbf{RichpediaMEL}} & \multicolumn{4}{c}{\textbf{WikiDiverse}}\\
\hline
\multicolumn{1}{c|}{\multirow{1}{*}{\textbf{Metrics}}}& \textbf{MRR} & \textbf{Hits@1} & \textbf{Hits@3} & \multicolumn{1}{c|}{\textbf{Hits@5}} & \textbf{MRR} & \textbf{Hits@1} & \textbf{Hits@3} & \multicolumn{1}{c|}{\textbf{Hits@5}} & \textbf{MRR}  & \textbf{Hits@1} & \textbf{Hits@3} & \textbf{Hits@5} \\
\hline
\multicolumn{1}{c|}{\texttt{LLaMA2-7B}}   &90.72  &86.71  &93.85  &\multicolumn{1}{c|}{95.65}   &83.48 &76.22  &89.61  &\multicolumn{1}{c|}{92.39}  &74.34  &64.87  &81.18  &85.85 \\ 
\multicolumn{1}{c|}{\texttt{LLaMA3.1-8B}}   &90.84  &86.67  &94.23  &\multicolumn{1}{c|}{96.09}   &88.05 &83.30  &91.83  &\multicolumn{1}{c|}{93.94}  &82.73  &75.79  &88.07  &91.39 \\ 
\hline
\multicolumn{1}{c|}{MMoE}   &\cellcolor{magenta!30}\textbf{93.75}  &\cellcolor{magenta!30}\textbf{90.77}  &\cellcolor{magenta!30}\textbf{96.27}  &\multicolumn{1}{c|}{\cellcolor{magenta!30}\textbf{97.41}}   &\cellcolor{magenta!30}\textbf{89.86} &\cellcolor{magenta!30}\textbf{85.51}  &\cellcolor{magenta!30}\textbf{93.43}  &\multicolumn{1}{c|}{\cellcolor{magenta!30}\textbf{95.31}}  &\cellcolor{magenta!30}\textbf{84.23}  &\cellcolor{magenta!30}\textbf{77.57}  &\cellcolor{magenta!30}\textbf{89.12}  &\cellcolor{magenta!30}\textbf{92.49} \\
\hline
\end{tabular*}
\caption{Evaluation of different LLMs of DME module on three MEL datasets.}
\label{table_different_llms}
\end{table*}

\subsection*{C\,\,\,Additional Results}
\hypertarget{additional_results}{}

\smallskip
\noindent\textbf{$\triangleright$ Different LLMs of DME module.}
\changed{
In the DME module, we use by default \texttt{GPT-3.5} as the ranking model for entity descriptions to identify the most contextually appropriate descriptions. To further investigate the ability of other open-source large language models to understand the relationship between entity description knowledge and mention context, we replace \texttt{GPT-3.5} with \texttt{LLaMA2-7B}~\citep{Hugo_2023a} and \texttt{LLaMA3.1-8B}~\citep{MetaAI_2024}. The corresponding experimental results are presented in Table~\ref{table_different_llms}. We can observe that the choice of ranking model has a significant impact on performance. Compared to \texttt{GPT-3.5}, both \texttt{LLaMA2} and \texttt{LLaMA3.1} perform slightly worse, with \texttt{LLaMA3.1} outperforming \texttt{LLaMA2}. This outcome aligns with expectations, as \texttt{GPT-3.5} consistently demonstrates superior performance across many task scenarios~\citep{Flavio_2024, Ujjwala_2024}.}